\documentclass{article}

\usepackage{arxiv}

\usepackage[utf8]{inputenc} 
\usepackage[T1]{fontenc}    
\usepackage{hyperref}       
\usepackage{url}            
\usepackage{booktabs}       
\usepackage{amsfonts}       
\usepackage{nicefrac}       
\usepackage{microtype}      
\usepackage{amsmath}
\usepackage{cleveref}       
\usepackage{lipsum}         
\usepackage{graphicx}
\usepackage{natbib}
\usepackage{doi}
\usepackage{adjustbox}
\usepackage{multirow}
\usepackage{multicol}
\usepackage{xcolor}
\usepackage{authblk}

\title{Automatic and standardized surgical reporting for central nervous system tumors}

\author[1,*]{David Bouget}
\author[1]{Mathilde Gajda Faanes}
\author[2,3]{Asgeir Store Jakola}
\author[4,5]{Frederik Barkhof}
\author[6]{Hilko Ardon}
\author[7]{Lorenzo Bello}
\author[8]{Mitchel S. Berger}
\author[8]{Shawn L. Hervey-Jumper}
\author[9,10]{Julia Furtner}
\author[11]{Albert J. S. Idema}
\author[12]{Barbara Kiesel}
\author[12]{Georg Widhalm}
\author[13]{Rishi Nandoe Tewarie}
\author[14]{Emmanuel Mandonnet}
\author[15]{Pierre A. Robe}
\author[16]{Michiel Wagemakers}
\author[17,18]{Timothy R. Smith}
\author[19,20]{Philip C. De Witt Hamer}
\author[21,22]{Ole solheim}
\author[1,23]{Ingerid Reinertsen}
\affil[1]{Department of Health Research, SINTEF Digital, Trondheim, Norway}
\affil[2]{Institute of Neuroscience and Physiology, Department of Clinical Neuroscience\\, University of Gothenburg, Gothenburg, Sweden}
\affil[3]{Department of Neurosurgery, Sahlgrenska University Hospital, Gothenburg, Sweden}
\affil[4]{Department of Radiology and Nuclear Medicine, Amsterdam University Medical Centers, Vrije Universiteit, Amsterdam, The Netherlands}
\affil[5]{Institutes of Neurology and Healthcare Engineering, University College London, London, United Kingdom}
\affil[6]{Department of Neurosurgery, Elisabeth-TweeSteden Hospital, Tilburg, The Netherlands}
\affil[7]{Neurosurgical Oncology Unit, Department of Oncology and Hemato-oncology, Humanitas Research Hospital, Milano, Italy}
\affil[8]{Department of Neurological Surgery, University of California, San Francisco, USA}
\affil[9]{Department of Biomedical Imaging and Image-guided Therapy, Medical University Vienna, Wien, Austria}
\affil[10]{Research Center for Medical Image Analysis and Artificial Intelligence, Faculty of Medicine and Dentistry, Krems, Austria}
\affil[11]{Department of Neurosurgery, Northwest Clinics,Alkmaar, The Netherlands}
\affil[12]{Department of Neurosurgery, Medical University Vienna, Wien, Austria}
\affil[13]{Department of Neurosurgery, Haaglanden Medical Center, The Hague, The Netherlands}
\affil[14]{Department of Neurological Surgery, H{\^o}pital Lariboisi{\`e}re, Paris, France}
\affil[15]{Department of Neurology and Neurosurgery, University Medical Center Utrecht, Utrecht, The Netherlands}
\affil[16]{Department of Neurosurgery, University Medical Center Groningen, University of Groningen, Groningen, The Netherlands}
\affil[17]{Computational Neuroscience Outcomes Center, Department of Neurosurgery, Brigham and Women’s Hospital, Boston, USA}
\affil[18]{Harvard Medical School, Boston, USA}
\affil[19]{Cancer Center Amsterdam, Brain Tumor Center, Amsterdam University Medical Centers, Amsterdam, The Netherlands}
\affil[20]{Department of Neurosurgery, Amsterdam University Medical Centers, Vrije Universiteit, Amsterdam, The Netherlands}
\affil[21]{Department of NeuroMedicine and Movement Science, NTNU, Trondheim, Norway}
\affil[22]{Department of Neurosurgery, St. Olavs hospital, Trondheim University Hospital, Trondheim, Norway}
\affil[23]{Department of Circulation and Medical Imaging, NTNU, Trondheim, Norway}
\affil[*]{david.bouget@sintef.no}

\hypersetup{
pdftitle={A template for the arxiv style},
pdfsubject={q-bio.NC, q-bio.QM},
pdfauthor={David S.~Hippocampus, Elias D.~Striatum},
pdfkeywords={First keyword, Second keyword, More},
}

\begin{document}
\maketitle

\begin{abstract}
Magnetic resonance (MR) imaging is essential for evaluating central nervous system (CNS) tumors, guiding surgical planning, treatment decisions, and assessing postoperative outcomes and complication risks. While recent work has advanced automated tumor segmentation and report generation, most efforts have focused on preoperative data, with limited attention to postoperative imaging analysis.
This study introduces a comprehensive pipeline for standardized postsurtical reporting in CNS tumors. Using the Attention U-Net architecture, segmentation models were trained, independently targeting the preoperative tumor core, non-enhancing tumor core, postoperative contrast-enhancing residual tumor, and resection cavity. In the process, the influence of varying MR sequence combinations was assessed. Additionally,
MR sequence classification and tumor type identification for contrast-enhancing lesions were explored using the DenseNet architecture. The models were integrated seamlessly into an automated and standardized reporting pipeline, following the RANO 2.0 guidelines. Training was conducted on multicentric datasets comprising $2\,000$ to $7\,000$ patients, incorporating both private and public data, using a 5-fold cross-validation. Evaluation included patient-, voxel-, and object-wise metrics, with benchmarking against the latest BraTS challenge results.
The segmentation models achieved average voxel-wise Dice scores of $87\%$, $66\%$, $70\%$, and $77\%$ for the tumor core, non-enhancing tumor core, contrast-enhancing residual tumor, and resection cavity, respectively. Classification models reached $99.5\%$ balanced accuracy in MR sequence classification and $80\%$ in tumor type classification.
The pipeline presented in this study enables robust, automated segmentation, MR sequence classification, and standardized report generation aligned with RANO 2.0 guidelines, enhancing postoperative evaluation and clinical decision-making.
The proposed models and methods were integrated into Raidionics, open-source software platform for CNS tumor analysis, now including a dedicated module for postsurgical analysis.
\end{abstract}


\section{Introduction}
Brain tumors encompass a diverse range of neoplasms with highly variable prognoses, ranging from benign to highly aggressive forms. The World Health Organization (WHO) currently classifies over $100$ subtypes based on molecular and histological profiles~\cite{louis20212021}. While some patients experience prolonged survival, many face rapid neurological and cognitive decline~\cite{day2016neurocognitive}. Accurate tumor characterization is crucial for prognosis, treatment planning, and surgical decision-making. However, the inherent biological complexity of CNS tumors  presents significant challenges.
CNS tumors are broadly categorized into primary or secondary types. Primary tumors, such as gliomas and meningiomas, originate within the brain or its supporting tissues. In particular, gliomas include the most aggressive and treatment-resistant forms (e.g., glioblastoma), as well as slowly progressive yet highly infiltrative forms ultimately undergoing malignant transformation (e.g., diffuse lower-grade)~\cite{chen2017glioma, jaeckle2011transformation}. On the other hand, secondary tumors result from metastatic spread to the brain.
Magnetic Resonance (MR) imaging is essential for tumor diagnosis, prognosis estimation, and treatment planning. Imaging-derived features such as volume, location, and structural involvement are critical for determining treatment planning, guiding surgical resection, and estimating postoperative risks~\cite{kickingereder2016radiomic}. Beyond clinical decision-making, these features are key components in developing predictive models for clinical research and personalized care~\cite{mathiesen2011two,sawaya1998neurosurgical,zinn2013extent}.
Despite significant advances in imaging technology, tumor characterization remains largely subjective. Manual tumor segmentation, the current gold standard for delineating tumor boundaries, remains labor-intensive and prone to intra- and inter-rater variability~\cite{binaghi2016collection}. Hence, it is rarely performed in routine practice due to time constraints. Instead, tumor attributes are often estimated visually or using crude methods (e.g., eyeballing or short-axis diameter measurements), introducing significant inconsistencies and reducing clinical utility~\cite{berntsen2020volumetric}. The lack of automated and standardized segmentation limits the integration of imaging-based biomarkers into clinical workflows. Robust computational methods are needed to improve precision, reproducibility, and efficiency in order to bridge the gap between imaging, molecular diagnostics, and personalized treatment strategies.

Post-operative MR imaging is critical for evaluating surgical outcomes, planning adjuvant therapy, and monitoring disease progression. However, altered anatomy, resection cavities, and postoperative blood products greatly complicate segmentation. In addition, residual tumor is often small and very fragmented, as opposed to tumor core appearance pre-operatively. The MICCAI BraTS challenge has been instrumental in advancing preoperative brain tumor segmentation for over a decade. In 2024, it was extended to include postoperative segmentation~\cite{de20242024}, introducing annotations for the enhancing tissue (ET), non-enhancing tumor core (NETC), surrounding non-enhancing FLAIR hyperintensity (SNFH), and resection cavity (RC). Top-performing methods employed CNN and Transformer-based variants (e.g., nnU-Net~\cite{isensee2021nnu}, Swin U-Netr~\cite{hatamizadeh2021swin}), often using ensemble techniques (e.g., STAPLE~\cite{warfield2004simultaneous}) and synthetic data augmentation~\cite{ferreira2024we}. Nonetheless, performance remains significantly lower than in the preoperative setting, with lesion-wise Dice scores reaching up to 78\% for NETC, 76\% for ET, and 71\% for RC. Other studies focusing on residual tumor segmentation attempted training from scratch~\cite{helland2023segmentation}, through active learning~\cite{luque2024standardized}, or transfer learning from a preoperative model~\cite{ghaffari2022automated}. In those studies, leveraging local datasets not publicly available, lower Dice scores were reported, at around 50\%-60\% Dice for ET.  A rather large difference compared to typical BraTS challenge results is shown, underscoring greater image and annotation variability in the postoperative setting.
Postprocessing techniques are frequently employed to remove artifacts and enforce anatomical consistency (e.g., ensuring NETC lies within TC).
In the current literature, segmenting the resection cavity has been relatively underexplored. For glioblastoma patients in a radiotherapy setting, using a dataset of 30 patients only, performance reached slightly over 80\% Dice score using a DenseNet variation~\cite{ermics2020fully}. Similar performances were obtained from Canalini et al. using a 3D U-Net, but leveraging longitudinal acquisitions~\cite{canalini2021comparison}, using a dataset of 50 patients. Alternatively, resection cavities were simulated before being segmented using self-supervised and semi-supervised learning~\cite{perez2021self}. Overall, none of the postoperative segmentation tasks can be considered solved, and further research is needed.

Beyond segmentation, standardized imaging reports are essential. The absence of structured and standardized surveillance reporting systems for glioma patients was already identified as a key limitation in 2018~\cite{weinberg2018management}. Although structured assessment has the potential to improve patient care and enhance both communication and decision-making, few solutions have been proposed. A preoperative standardized reporting based on MR scans was proposed as part of the Raidionics platform~\cite{bouget2023raidionics}, assembled from features extracted from automatic segmentation and covering the main CNS tumor types (i.e., gliomas, meningiomas, and metastases). Regarding postoperative assessment and surveillance in glioma patients~\cite{weinberg2018management, parillo2024brain}, both works focused on designing new scoring methods and structured assessment, however not automatically performed from the analysis of MR scans.
With the increased interest in postoperative CNS tumor segmentation, robust and state-of-the-art models for ET and RC are emerging. As such, there is a clear opportunity for advancement in readily available solution for postoperative standardized and automatic reporting.

This work introduces a comprehensive pipeline for automated postsurgical assessment in CNS tumors, with the following contributions: (i) development of robust segmentation models for tumor core, non-enhancing tumor core, residual tumor, and resection cavity, (ii) thorough validation and benchmarking against BraTS, including ablation studies on required MR sequences, (iii) extension and refinement of image-based standardized reporting in line with RANO 2.0 guidelines, and (iv) integration into the Raidionics software, providing access to the latest segmentation, classification, and reporting methods.

\section{Data}
\label{sec:Data}

\begin{figure}[!ht]
\centering
\includegraphics[scale=0.70]{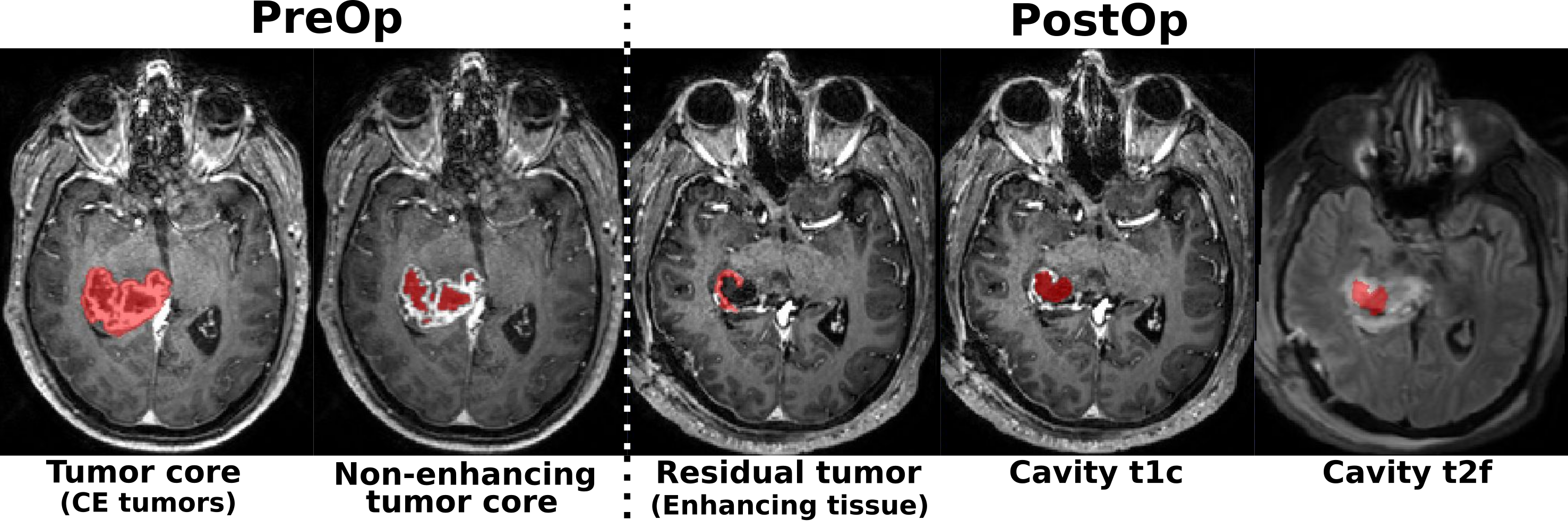}
\caption{Illustration of the different structures targeted for model segmentation training: tumor core (for contrast-enhancing tumors), non-enhancing tumor core, postoperative residual tumor (enhancing tissue), resection cavity in t1c MR scan, and resection cavity in t2f MR scans (from left to right).}
\label{fig:dataset-illu}
\end{figure}

All data used in this study were obtained from a previously described private dataset~\cite{bouget2022preoperative}, and the publicly available datasets from the BraTS challenges~\cite{karargyris2023federated,moawad2024brain,de20242024,labella2024brain}.
Given the diverse segmentation and classification tasks addressed in this study, different specific subsets were compiled, as detailed below. An overview of the segmentation subsets is provided in Table~\ref{tab:datasets-overview}.
First, for the MR sequence classification task, a total of $1\,000$ MR scans were randomly selected for each combination of MR sequence type and acquisition timestamp (i.e., preoperatively and postoperatively), resulting in a final subset of $8\,000$ MR scans. The MR sequence types considered are:  gadolinum-enhanced T1-weighted (noted t1c), T1-weighted (noted t1w), FLAIR (noted t2f), and T2-weighted (noted t2w). 
Second, for the contrast-enhancing tumor type classification, a total of $500$ pre-operative t1c MR scans were randomly selected for each class (i.e., glioblastoma, meningioma, and metastasis), resulting in a final subset of $1\,500$ MR scans.
Finally, for the segmentation task, four distinct subsets were assembled using a mixture of patient data at different timepoints. Each subset targets a specific structure to segment, the tumor core for contrast-enhancing tumors (i.e., TC), the non-enhancing tumor core (i.e., NETC), the residual tumor (or enhancing tissue), and finally the resection cavity (as illustrated in Fig.~\ref{fig:dataset-illu}. In table~\ref{tab:datasets-overview}, the total number of unique patients, as well as the total number of investigation from different timestamps, are summarized. Additionally, the incremental decrease of number of investigations, arising from the inclusion of additional MR sequences, is also reported. Different cut-off values were applied to determine whether the segmentation target was sufficiently visible in a given MR scan to be considered a positive sample: $0.175$\,ml for postoperative residual tumor~\cite{wen2023rano}, $0.05$\,ml for NETC, $0.1$\,ml for resection cavity, and $0.1$\,ml for TC.
For the private data, tumor core, residual tumor, and resection cavities were manually segmented in 3D by trained raters. On preoperative t1c scans, the tumor core region was defined as gadolinum-enhancing tissue, necrosis, and cysts. On postoperative t1c scans, the enhancing tissue region was defined as gadolinum-enhancing tissue. Finally, the resection cavity was defined as regions with a signal isointense to cerebrospinal fluid, potentially including air, blood, or proteinaceous materials when recent.
A more detailed description of each cohort and dataset is available in Section 1 of the Supplementary Material.

\begin{table}[!ht]
\caption{Overview of the datasets used for the different segmentation tasks. A - Preoperative tumor core (for contrast-enhancing CNS tumor), B - Pre/postoperative non-enhancing tumor core (NETC), C - Postoperative residual tumor, D - Resection cavity. (*) For 306 lower-grade glioma patients, the resection cavity is only available in t2f MR scans. (+) The BraTS challenge data is considered as a single data source even if collected from multiple hospitals.}
\adjustbox{max width=\textwidth}{
\begin{tabular}{lc|cccc|cccc|c}
Name & Target & Timestamps & Patients & Sources & Positives & t1c & +t1w & +t2f & +t2w & Volume (ml) \tabularnewline
\hline
A & Tumor core& $7\,171$ & $7\,171$ & $17$ & $7\,106$ & $7\,171$ & - & - & - & $30.28\pm27.95$\tabularnewline
B & NETC & $4\,427$ & $3\,724$ & $1^{+}$ & $2\,507$ & $4\,427$ & $4\,427$ & $4\,427$ & $4\,427$ & $05.43\pm13.83$\tabularnewline
C & Residual tumor & $2\,648$ & $1\,945$ & $16$ & $1\,674$ & $2\,648$ & $2\,616$ & $2\,463$ & $2\,224$ & $04.98\pm06.77$\tabularnewline
D & Resection cavity & $2\,275$ & $1\,572$ & $16$ & $2\,048$ & $1\,969$ & $1\,926$ & $1\,859$ ($306$*) & $1\,826$ & $15.95\pm19.69$\tabularnewline
\end{tabular}
}
\label{tab:datasets-overview}
\end{table}

\section{Methods}

\subsection{Segmentation models training}
\label{subsec:training}
\begin{figure}[!ht]
\centering
\includegraphics[scale=0.6]{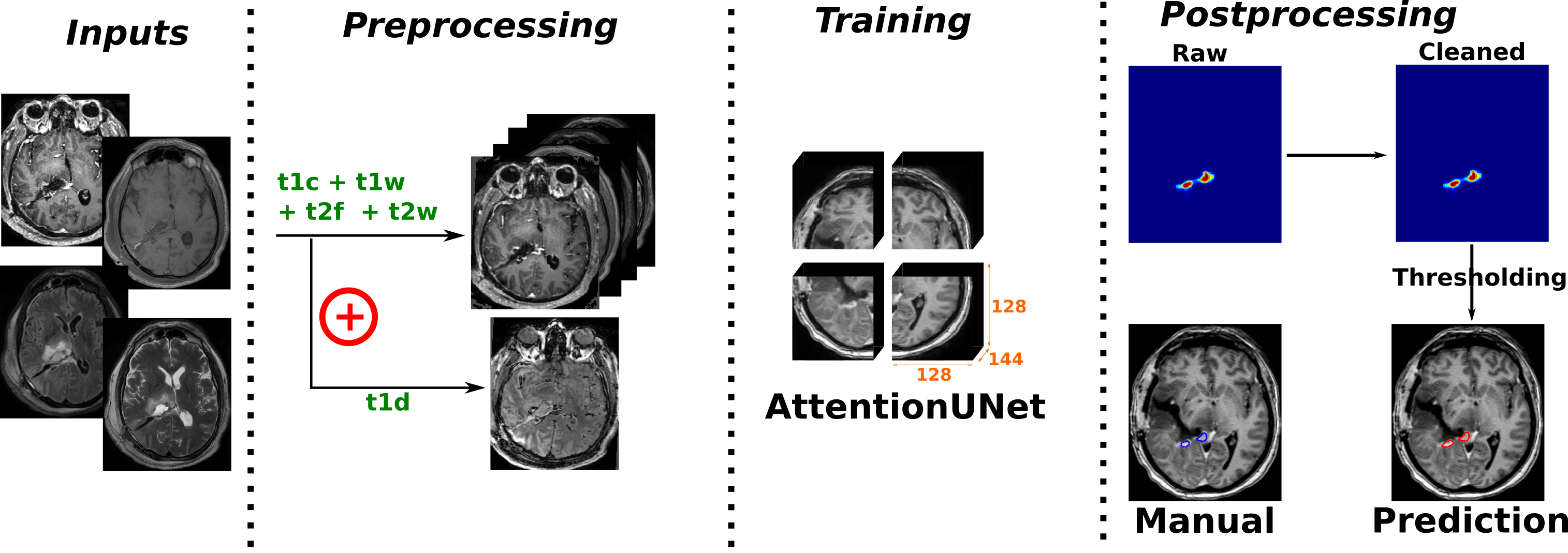}
\caption{Illustration of the segmentation model training pipeline including the following four steps: MR sequence input selection, preprocessing, training using the Attention U-Net architecture, and finally postprocessing.}
\label{fig:segmentation-pipeline-illu}
\end{figure}

Segmentation models were trained following the pipeline illustrated in Fig.~\ref{fig:segmentation-pipeline-illu}, with preprocessing variations based on the available input MR sequences. Each step of the pipeline is further detailed in the following paragraphs.

\subsubsection{Input selection and preprocessing}
Given the possibility for one or multiple MR scans to be available for any given patient, incremental combinations of input MR sequences are made possible. Additionally, pairs of input sequences can be subtracted to generate new input channels, a technique inspired by the best-performing team for meningioma segmentation in the BraTS 2024 challenge. The inclusion of such subtraction-based inputs is indicated by an additional marker \textit{(d)} within the experiment name (e.g., \textit{t1d} for subtraction between t1c and t1w scans).

The following preprocessing steps were applied in the specified order: (i) resampling to an isotropic voxel spacing of $1\,\text{mm}^3$ using first-order spline interpolation, (ii) tight cropping around the patient's head to exclude the background, (iii) subtraction of two input sequences to create a new difference input channel (when applicable), (iv) intensity clipping within the range $[0, 99.5]\%$, and (v) zero-mean normalization of all nonzero values.
    
\subsubsection{Architecture design and training strategy}
The Attention U-Net architecture~\cite{oktay2018attention} has been used in this work with the following specifications: $5$ levels, filter sizes of $[16, 64, 128, 256, 512]$, instance normalization, a dropout rate of $0.2$, and an input size of $128\times128\times144$\,voxels.

The loss function was the combination of Dice and Cross-Entropy, with a sigmoid final activation for single-class segmentation tasks (i.e., including the background). Model training was conducted using the AdamW optimizer with an initial learning rate of $5e-4$, combined with an annealing scheduler. Training was performed over $800$ epochs with a batch size of $16$ elements, using a $2$ step gradient accumulation strategy to achieve an effective batch size of $32$ elements.

For data augmentation, a random crop of $128\times128\times144$\,voxels was applied to each input sample. Subsequently, a random combination of geometric, each with a $50\%$ probability to happen over any given axis, and intensity-based transformations with a $50\%$ probability were performed. The geometric transformations include flipping, rotation within the range $[-20^{\circ}, 20^{\circ}]$, translation of up to $20\%$ of the axis dimension, and zoom scaling of up to $15\%$. The intensity-based transformations include intensity scaling and shifting (up to $10\%$), gamma contrast adjustments in the range $[0.5, 2.0]$, Gaussian noise addition, and patch dropout or inversion with patch sizes of $10\times10\times10$\,voxels and up to $75$ elements.

\subsubsection{Inference and postprocessing}
For the inference step, a sliding-window approach with $50\%$ overlap between consecutive patches along each spatial dimension was performed. Unlike the patch size employed during training, the inference patch size was set to $160\times160\times160$\,voxels. During the inference process, no data augmentation was performed over the input samples.
Subsequently, a two-step postprocessing refinement was designed to clean the predictions. With the first step, only the prediction probabilities lying inside a binary mask of the brain location were kept. Second, noise in prediction probabilities was removed by identifying connected component predictions with an area lower than $0.05\,ml^{3}$ or not visible in consecutive 2D slices.

\subsection{Single timepoint and surgical standardized reporting}
\begin{figure}[!ht]
\centering
\includegraphics[scale=0.5]{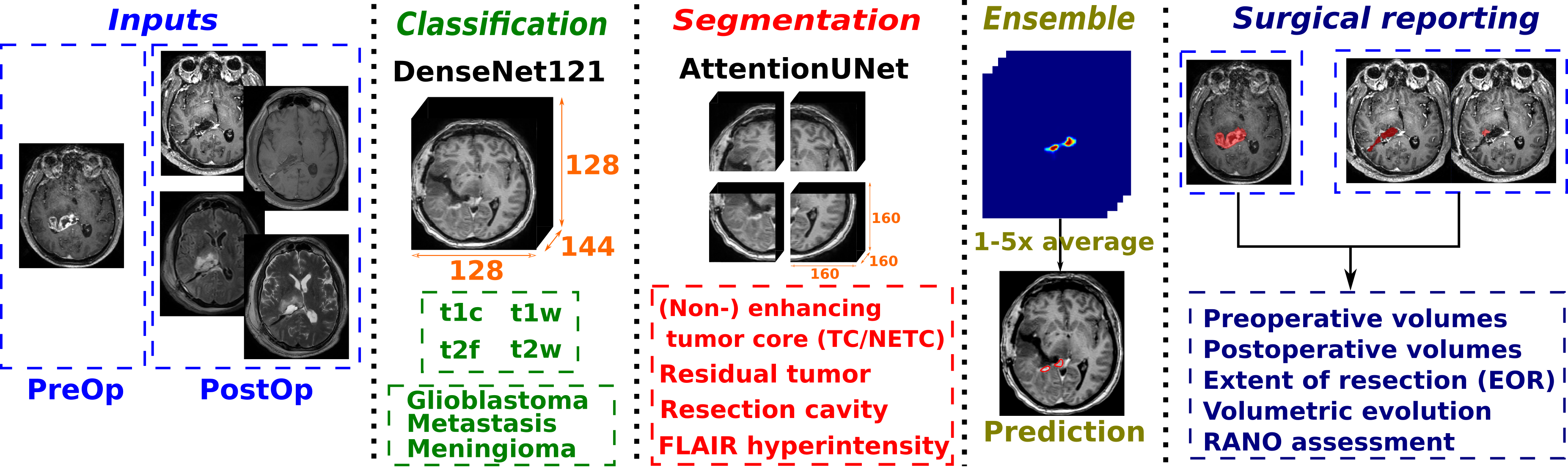}
\caption{Pipeline illustration for surgical reporting using any given set of input MR scans. The different steps are: MR sequence classification, contrast-enhancing tumor type classification, input-agnostic structures segmentation, model ensembling, and finally reporting generation (per timepoint and postsurgical).}
\label{fig:inference-pipeline-illu}
\end{figure}

The proposed pipeline for standardized surgical reporting, illustrated in Fig.~\ref{fig:inference-pipeline-illu}, starts with a classification step to automatically identify the MR sequence for all provided input scans and the type of contrast-enhancing tumor (if applicable). Then, segmentation of multiple structures is performed using the latest models, with the possibility to include an extra step of ensembling to generate more robust results. Finally, a surgical reporting can be generated, complementing the standardized reporting computed for each timepoint. Each step is described in-depth in the remainder of the section.

\subsubsection{MR scan sequence and contrast-enhancing tumor type classification}
In order to automatically assign the proper MR sequence to each input scan and identify the contrast-enhancing tumor type, 3D classification model training was performed. The DenseNet121 architecture~\cite{huang2017densely} has been used with $64$ filters in the first convolution layer, a growth rate of $32$, batch normalization, and an input size of $128\times128\times144$\,voxels.

The loss function was Cross-Entropy and the multi-class accuracy was employed as metric. Model training was conducted using the AdamW optimizer with an initial learning rate of $5e-4$, combined with an annealing scheduler. Training was performed over $800$ epochs with a batch size of $8$ elements, using a $4$ step gradient accumulation strategy to achieve an effective batch size of $32$ elements. The same data augmentation techniques as described in the previous section were used.

\subsubsection{Segmentation and model ensembling}
The inference process is the one previously described and by default no runtime data augmentation nor model ensembling is performed. Both options can be enabled by the user in order to improve the robustness of the segmentation results only at the expense of longer processing time. The available runtime data augmentation techniques, using the same parameters as during training, include axis flipping, rotation, and gamma contrast. For model ensembling, the segmentation results from one to five models can be combined by returning either the mean probability (i.e., \textit{average} option) or the maximum probability (i.e., \textit{amax} option) for each voxel.

\subsubsection{Global structures' segmentation refinement}
As each structure segmentation model was trained independently, a refinement step leveraging global context is appropriate to ensure consistency across all structures. In addition to the models described in this study, the FLAIR hyperintensity (i.e., SNFH) segmentation models, trained using the training protocols~\cite{Mathilde2025}, were included in this step.

For contrast-enhancing tumors: Preoperatively, the tumor core mask is kept unchanged and used as reference. The NETC mask is refined by retaining only regions that overlap with the tumor core mask, and the SNFH mask is modified by subtracting the tumor core region. The non-overlapping tumor core and SNFH masks can be combined to form the whole tumor mask. Postoperatively, the residual tumor mask is kept unchanged and used as reference. The resection cavity and enhancing tissue masks are subtracted from the SNFH mask.

For non contrast-enhancing tumors: No preoperative tumor core or postoperative residual tumor masks are available. In both preoperative and postoperative settings, the SNFH mask serves as the whole tumor mask. Specifically postoperatively, the resection cavity mask is subtracted from the SNFH mask.

\subsubsection{Standardized reporting}
Standardized reports for single timepoints (i.e., preoperative and postoperative) are computed in the same way as described in our previous study~\cite{bouget2022preoperative}. The major variation comes from the selection of segmentation models used for the different use-cases, when only a single tumor model was available before. For contrast-enhancing tumors, the tumor core segmentation model operating over gadolinum-enhanced T1-weighted MR scans is used preoperatively and the residual tumor segmentation model is used postoperatively over up to four MR sequences. For non contrast-enhancing tumors, the unified SNFH segmentation model is used~\cite{Mathilde2025}. For all tumor types, the unified resection cavity and necrosis segmentation models are used over up to four MR sequences.
The set of computed features has been extended to include diameter characteristics (i.e., long-axis, short-axis, Feret, and equivalent area), tumor-to-brain ratio, and the Brain-Grid classification system for cerebral gliomas~\cite{latini2019novel}.

The standardized surgical report features the same distinction between contrast-enhancing and non contrast-enhancing tumors. Preoperative and postoperative volumes for all segmented structures are reported, when applicable. In addition, postsurgical volumetric evolution percentages are reported for each segmented structure, which is the equivalent to the extent-of-resection when the considered structure is the tumor. Finally, the overall surgical assessment, i.e., complete, near total, or subtotal resection, is provided to complement the standardized surgical report, following the latest RANO guidelines~\cite{wen2023rano}.

\section*{Validation studies}
In this work, the focus lies primarily on assessing segmentation and classification models' performance, and as such four experiments were conducted. A single training protocol has been followed for all presented models, namely a 5-fold cross-validation. All datasets were randomly split into $5$ folds, simply enforcing for a single patient's data not to be featured in multiple folds when MR scans were available at multiple timestamps. During training, $3$ folds were used as training set, one fold as validation set, and the remaining fold as test set, following an iterative process. For the validation studies, both postprocessing steps were used for the residual tumor structure, and only the first step was applied for all other structures.

\subsection{Metrics}
\label{subsec:measandmet}
For quantifying and comparing models' performances, the following patient-wise, voxel-wise, and object-wise metrics were computed. Each metric was computed between the ground truth volume and a binary representation of the probability map generated by a trained model. The binary representation is computed for ten different equally-spaced probability thresholds in the range $[0, 1]$. For the probability threshold providing the best results, pooled estimates computed from each fold's results are then reported for each metric~\cite{killeen2005alternative} to provide overall results, reported as mean and standard deviation (indicated by $\pm$) in the tables. Voxel-wise and object-wise results are only reported for the positive samples following the definition provided below.

\paragraph{Patient-wise:} Patient-wise metrics assess the classification ability of a given segmentation or classification model. For segmentation models, the cut-off volume values presented in the Data section were used to determine positive cases (i.e., including the structure of interest) from negative cases (i.e., not include the structure of interest). Furthermore, a voxel-wise Dice overlap of simply $0.1\%$ between model prediction and ground truth was required for a positive case to be considered true positive (TP). From the identification of true/false positive/negative samples at a patient level, the following metrics were then computed: recall, precision, specificity, and balanced accuracy (noted bAcc). 

\paragraph{Voxel-wise:} Voxel-wise metrics assess the ability of a segmentation model by considering each voxel independently. The following metrics were computed between the ground truth volume and the binary model prediction: Dice score, recall, precision, and 95th percentile Hausdorff distance (noted HD95).

\paragraph{Object-wise:} Object-wise metrics assess the ability of a segmentation model to detect all components of the structure to segment (i.e., multiple tumor components). The trained models not being instance segmentation models, a connected components approach coupled to a pairing strategy was employed to associate ground truth and model predictions' components. A strict assignment was performed using the Hungarian algorithm. A minimum component size of $75$\,voxels (down to $50$\,voxels for the NETC structure) was enforced and objects below the threshold were discarded. Dice score, recall, precision, and 95th percentile Hausdorff distance were computed.

\subsection{Experiments}
First, (i) classification performance analyses for MR sequence and tumor type identification were executed. Next, (ii) a performance analysis of the contrast-enhancing tumor core segmentation model and NETC segmentation model was performed. Then, (iii) a postoperative segmentation performance experiment was conducted to identify the impact of using a varying number of MR sequences as input and present the best performing models for the enhancing tissue and resection cavity categories. Finally, (iv) a detailed analysis was carried out to highlight performance differences across the different cohorts in our dataset, as well as in comparison with the BraTS challenge for external benchmarking.

\section{Results}
All experiments were performed using computers with the following specifications: Intel Core Processor (Broadwell, no TSX, IBRS) CPU with $16$ cores, $64$GB of RAM, Tesla A40 ($46$ GB) or A100 ($80$GB) dedicated GPUs, and NVMe hard-drives. Training and inference processes were implemented in Python $3.11$ using \texttt{PyTorch} v2.4.1, \texttt{PyTorch Lightning} v2.4.0 , and \texttt{MONAI} v1.4.0~\cite{cardoso2022monai}. The source code used for computing the metrics and performing the validation studies, all trained models, and inference code, are publicly available through our Raidionics platform~\cite{bouget2023raidionics}.

\subsection{(i) Classification performances}
Classification performances for both MR sequence and tumor type tasks are summarized in Table.~\ref{tab:classif-results}. From the use of a large dataset of $8\,000$ samples, MR sequence classification performances are almost perfect with a 99\% balanced accuracy score. On the other hand, classification performances for the contrast-enhancing tumor type are only reaching 85\% balanced accuracy. The results can be explained by the use of a relatively smaller dataset, when compared to the MR sequence classification dataset, and by the more difficult nature of the task when only leveraging the t1c MR scans.

\begin{table}[!h]
\centering
\caption{Multiclass-averaged classification performance for distinguishing between MR sequence type (Sequence) and between contrast-enhancing tumor type (Type).}
\adjustbox{max width=\textwidth}{
\begin{tabular}{c|ccccccc|}
Task & Recall & Precision & Specificity & F1-score & Accuracy & bACC\tabularnewline
\hline
Sequence & $98.83\pm0.19$ & $98.84\pm0.19$ & $99.61\pm0.06$ & $98.83\pm0.19$ & $99.42\pm0.09$ & $99.22\pm0.13$ \tabularnewline
Type & $80.06\pm2.27$ & $80.46\pm2.80$ & $89.99\pm1.23$ & $80.02\pm2.42$ & $86.67\pm1.61$ & $85.02\pm1.75$ \tabularnewline
\end{tabular}
}
\label{tab:classif-results}
\end{table}

\begin{figure}[!t]
\centering
\includegraphics[scale=0.45]{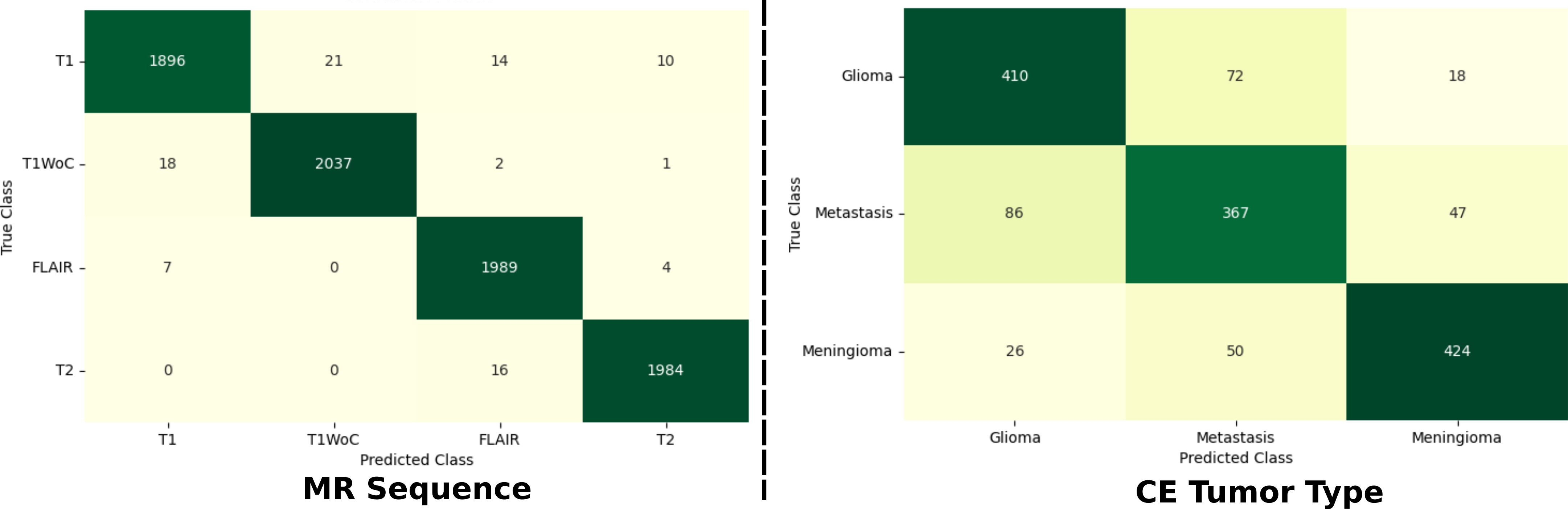}
\caption{Confusion matrices for both classification tasks where the predicted class is shown in the x-axis and the ground truth class in the y-axis.}
\label{fig:classif-confusion-matrices}
\end{figure}

\begin{figure}[!ht]
\centering
\includegraphics[scale=0.55]{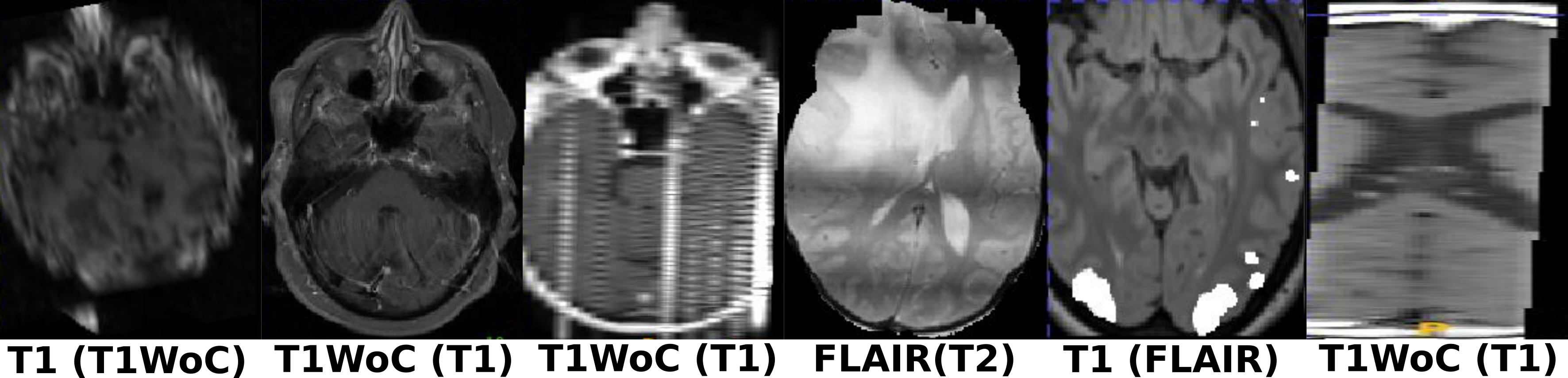}
\caption{Examples of misclassified MR scans due to extreme motion artifacts, reconstruction artifacts, noise, or cropping. The predicted MR sequence is indicated first and the ground truth MR sequence is specified in parenthesis.}
\label{fig:misclassification-sequence-illu}
\end{figure}

To further investigate where the classification models failed, confusion matrices were computed for each task (cf. Figure~\ref{fig:classif-confusion-matrices}). For the MR sequence task, frequent misclassifications occurred between gadolinum-enhanced T1-weighted and regular T1-weighted MR scans. Similarly, confusion was common between FLAIR and T2 MR sequences. These results are expected, as each of these pairs shares similar physical imaging characteristics. Upon detailed review of the 93 misclassified cases, it was found that in 61 instances the ground truth labels were incorrect, while the model predictions were actually accurate. Therefore, only 32 cases (i.e., $0.4\%$) were truly misclassified. These errors were primarily attributed to extreme cropping, imaging artifacts, noise, or blurriness (as illustrated in Fig.~\ref{fig:misclassification-sequence-illu}).
Regarding the tumor type task, both glioblastomas and meningiomas were usually correctly classified. The largest confusion came from the metastasis group, more often misclassified as either of the two other groups.

\subsection{(ii) Contrast-enhancing tumor core and NETC segmentation performances}
Tumor core segmentation performance in preoperative MR scans for contrast-enhancing tumors are reported in Table.~\ref{tab:seg-results-preop-tc}. Overall patient-wise metrics indicate an almost perfect detection rate with 99\% recall and precision. However, from the heavy imbalance between positive and negative samples in dataset A, the balanced accuracy and specificity were negatively impacted. Both voxel-wise and object-wise Dice scores reached high values above 85\%, indicating strong efficiency both on average and for each tumor type individually. The different metrics are quite homogeneous across the board. The lowest voxel-wise Dice score (85\%) was obtained for the metastasis group while the highest (89\%) was obtained for the glioma group. Supported by the average volumes reported in Table.S1, larger tumors exhibit higher Dice scores on average. Interestingly, the object-wise Dice score for the metastasis group is almost as high as for the other two groups. Despite metastasis being on average smaller and more fragmented than meningioma and glioma, the results showcase high and stable segmentation performance. A deeper performance investigation based on tumor type and cohort is available in Table.S5 and Fig.S1.

\begin{table}[h]
\caption{Preoperative contrast-enhancing tumor core segmentation performances, over t1c MR scans from dataset A. The tumor types are glioma (GLI), meningioma (MEN), and metastasis (MET).}
\adjustbox{max width=\textwidth}{
\begin{tabular}{rr||cccc||cccc||cccc}
 & & \multicolumn{4}{c||}{Patient-wise} & \multicolumn{4}{c||}{Voxel-wise} & \multicolumn{4}{c}{Object-wise} \tabularnewline
Type & \# Samples & Recall & Precision & Specificity & bAcc & Dice & Recall & Precision & HD95 & Dice & Recall & Precision & HD95\tabularnewline
GLI & $3\,655$ & $99.67\pm00.22$ & $99.81\pm00.11$ & $26.89\pm38.87$ & $63.28\pm19.38$ & $89.01\pm14.02$ & $89.75\pm13.82$ & $90.36\pm13.77$ & $4.29\pm10.19$ & $88.27\pm14.11$ & $88.93\pm14.42$ & $90.25\pm13.29$ & $2.88\pm3.75$\tabularnewline
MEN & $2\,278$ & $98.29\pm00.65$ & $99.91\pm00.17$ & $79.02\pm40.00$ & $88.65\pm19.89$ & $85.94\pm20.67$ & $85.61\pm21.84$ & $88.41\pm19.64$ & $7.93\pm20.55$ & $86.32\pm20.35$ & $85.74\pm21.74$ & $91.16\pm14.49$ & $2.42\pm3.92$ \tabularnewline
MET & $1\,214$ & $98.83\pm00.76$ & $99.16\pm00.44$ & $59.88\pm22.42$ & $79.36\pm10.99$ & $85.05\pm15.41$ & $83.93\pm16.71$ & $88.92\pm14.11$ & $10.36\pm21.72$ & $83.67\pm20.82$ & $86.29\pm14.91$ & $90.41\pm13.24$ & $1.49\pm1.30$ \tabularnewline
\hline
All & $7,171$ & $99.09\pm00.08$ & $99.73\pm00.07$ & $39.46\pm16.84$ & $69.27\pm08.39$ & $87.32\pm16.80$ & $87.44\pm17.43$ & $89.44\pm16.06$ & $6.49\pm16.55$ & $86.84\pm17.67$ & $87.46\pm17.27$ & $90.52\pm13.77$ & $2.52\pm3.66$\tabularnewline
\end{tabular}
}
\label{tab:seg-results-preop-tc}
\end{table}

Segmentation performances for the NETC category are reported in Table.~\ref{tab:seg-results-necrosis}. The incremental addition of MR scans as input seemed to have a marginal effect only on the average performances. Both average pixel-wise and object-wise Dice scores went from 65.6\% with one input sequence to 66.8\% with three. Given the higher negative-to-positive sample ratio in dataset B, patient-wise specificity and balanced accuracy are above 70\% and 80\% respectively. The final addition of t2f slightly worsened overall performance with Dice score values around 65.5\% yet providing a higher recall than using t1c sequences only. The very fragmented nature of the NETC structure is clearly visible from the large difference between pixel-wise and object-wise HD95 values. By removing all fragments below a predefined volume threshold, as conventionally done in the BraTS challenge, the distance between the largest paired components is drastically reduced.
A deeper performance investigation based on tumor type and NETC volume is available in Table.S6, Table.S7, and Fig.S2.

\begin{table}[h]
\caption{Non-enhancing tumor core segmentation performances over dataset B, for different sets of MR sequences as input.}
\adjustbox{max width=\textwidth}{
\begin{tabular}{r||cccc||cccc||cccc}
 & \multicolumn{4}{c||}{Patient-wise} & \multicolumn{4}{c||}{Voxel-wise} & \multicolumn{4}{c}{Object-wise} \tabularnewline
Inputs & Recall & Precision & Specificity & bAcc & Dice & Recall & Precision & HD95 & Dice & Recall & Precision & HD95\tabularnewline
\hline
t1c & $94.24\pm01.47$ & $82.20\pm01.15$ & $73.35\pm01.58$ & $83.80\pm00.92$ & $65.64\pm30.84$ & $70.78\pm33.30$ & $69.45\pm29.30$ & $10.21\pm16.88$ & $65.61\pm31.89$ & $75.03\pm32.69$ & $75.35\pm25.96$ & $3.75\pm4.70$\tabularnewline
+ t1wt1d & $94.99\pm01.19$ & $80.86\pm01.38$ & $70.65\pm01.71$ & $82.82\pm01.18$ & $66.11\pm30.47$ & $71.99\pm32.55$ & $68.90\pm29.05$ & $10.37\pm17.39$ & $66.27\pm31.23$ & $76.71\pm31.62$ & $74.21\pm26.40$ & $3.78\pm4.74$\tabularnewline
+ t2f & $95.08\pm01.53$ & $81.21\pm03.26$ & $71.17\pm05.16$ & $83.13\pm01.88$ & $66.72\pm30.02$ & $72.15\pm32.28$ & $69.93\pm28.59$ & $10.47\pm17.99$ & $66.85\pm31.22$ & $76.69\pm31.52$ & $75.03\pm25.99$ & $3.71\pm4.89$ \tabularnewline
+ t2w & $95.97\pm01.37$ & $77.57\pm02.12$ & $63.76\pm03.10$ & $79.87\pm00.93$ & $65.30\pm29.94$ & $75.22\pm31.55$ & $64.71\pm29.13$ & $10.78\pm18.62$ & $65.43\pm30.56$ & $79.35\pm30.78$ & $68.94\pm27.70$ & $3.61\pm4.37$\tabularnewline
\end{tabular}
}
\label{tab:seg-results-necrosis}
\end{table}

\subsection{(iii) Postoperative segmentation performances}
First, segmentation performances for the contrast-enhancing residual tumor category, in postoperative MR scans, are reported in Table~\ref{tab:gbm-postop-seg-results}. The best performances are achieved for the model trained using all four input MR sequences with a 70\% pixel-wise and object-wise average Dice score. An overall decrease of 7\% Dice can be noticed when training a model using a single MR sequence as input, and each input sequence iteratively boosts performances by 1-2\% Dice. Looking at the predictive performances, the model using only t1c inputs exhibits oversegmentation and struggles to properly identify patients with residual tumor from patients without. A substantial improvement in patient-wise specificity and balanced accuracy is obtained from using both t1c and t1w MR sequences as input with a respective 13\% and 6.5\% increase. Having access to information from two different but highly correlated MR sequences helped the model to better differentiate between residual tumor and blood products. However, the best patient-wise specificity score reached only 70\% with all four input MR sequences, highlighting the struggle not to oversegment given the very fragmented nature of residual tumor.

\begin{table}[!ht]
\caption{Overall segmentation performance summary for contrast-enhancing residual tumor over dataset C, with incremental inclusion of MR sequences as input.}
\adjustbox{max width=\textwidth}{
\begin{tabular}{r||cccc||cccc||cccc}
 & \multicolumn{4}{c||}{Patient-wise} & \multicolumn{4}{c||}{Voxel-wise} & \multicolumn{4}{c}{Object-wise} \tabularnewline
Inputs & Recall & Precision & Specificity & bAcc & Dice & Recall & Precision & HD95 & Dice & Recall & Precision & HD95\tabularnewline
\hline
t1c & $97.18\pm01.12$ & $77.01\pm01.03$ & $50.70\pm03.46$ & $73.94\pm01.28$ & $63.26\pm26.77$ & $70.14\pm27.26$ & $64.30\pm29.11$ & $12.85\pm21.27$ & $64.27\pm26.42$ & $71.25\pm27.99$ & $68.77\pm27.49$ & $5.52\pm6.18$ \tabularnewline
+ t1wt1d & $95.31\pm01.08$ & $81.37\pm04.29$ & $63.60\pm06.61$ & $79.46\pm03.01$ & $66.10\pm26.53$ & $70.58\pm27.80$ & $68.61\pm28.10$ & $11.14\pm17.31$ & $67.06\pm26.40$ & $72.57\pm27.76$ & $72.99\pm25.82$ & $4.71\pm5.27$ \tabularnewline
+ t2f & $95.17\pm01.32$ & $83.27\pm03.18$ & $68.06\pm07.85$ & $81.61\pm03.46$ & $67.72\pm26.36$ & $70.85\pm27.65$ & $70.86\pm27.70$ & $10.37\pm17.74$ & $68.10\pm26.55$ & $72.35\pm27.89$ & $74.91\pm25.16$ & $4.52\pm5.51$\tabularnewline
+ t2w & $95.27\pm00.75$ & $84.03\pm02.51$ & $70.61\pm04.78$ & $82.94\pm02.40$ & $70.24\pm25.97$ & $73.49\pm26.60$ & $73.06\pm27.18$ & $8.66\pm16.05$ & $70.42\pm26.02$ & $74.93\pm26.71$ & $77.17\pm23.81$ & $4.32\pm5.38$ \tabularnewline
\end{tabular}
}
\label{tab:gbm-postop-seg-results}
\end{table}

Segmentation performances for the resection cavity in postoperative MR scans are reported in Table~\ref{tab:cav-postop-seg-results}. The resection cavity often comprise a single region and as such object-wise metrics are closely matching the pixel-wise metrics. The lowest object-wise Dice score is obtained when using a single t2f MR scan as input, reaching only 66.5\%. The peculiarity of this use-case is the mixture of resection cavities in patients suffering from glioblastoma or diffuse lower-grade glioma, potentially rendering the task more difficult from higher brain structure and appearance variability. Regarding the sequential inclusion of MR sequence from t1w to t2w, performances are almost identical using a single input or the four MR sequences, hovering around 78\% object-wise Dice. From the patient-wise results, the classification ability is relatively poor with around 25\% specificity and 60\% balanced accuracy. As highlighted by the dataset distribution in Table. S4, the positive rate lies at 90\%. Hence, very few negative samples are provided to the model during training for mitigating such oversegmentation behaviour. Yet, the patient-wise precision is relatively high with 90\%, indicating that not many patients were misclassified regarding the presence or absence of a resection cavity. The specificity values are negatively impacted by the very shallow pool of negative samples in the dataset.

\begin{table}[h]
\caption{Overall segmentation performance summary for the resection cavity in dataset D, for different sets of MR sequences as input.}
\adjustbox{max width=\textwidth}{
\begin{tabular}{r||cccc||cccc||cccc}
 & \multicolumn{4}{c||}{Patient-wise} & \multicolumn{4}{c||}{Voxel-wise} & \multicolumn{4}{c}{Object-wise} \tabularnewline
Inputs & Recall & Precision & Specificity & bAcc & Dice & Recall & Precision & HD95 & Dice & Recall & Precision & HD95\tabularnewline
\hline
t2f &$92.50\pm01.65$ & $87.28\pm01.58$ & $11.95\pm03.57$ & $52.23\pm02.56$ & $65.16\pm26.04$ & $71.04\pm27.26$ & $66.57\pm25.77$ & $23.12\pm31.69$ & $66.55\pm25.84$ & $71.24\pm27.70$ & $68.91\pm26.32$ & $6.38\pm6.69$ \tabularnewline
t1c & $99.24\pm00.70$ & $89.83\pm02.09$ & $16.38\pm05.72$ & $57.81\pm02.92$ & $76.23\pm24.79$ & $79.81\pm24.40$ & $77.06\pm25.82$ & $13.86\pm26.75$ & $77.93\pm23.88$ & $80.01\pm24.75$ & $80.55\pm23.69$ & $4.54\pm5.49$\tabularnewline
+t1wt1d & $98.71\pm00.72$ & $90.84\pm01.80$ & $25.70\pm05.02$ & $62.20\pm02.25$ & $76.42\pm24.47$ & $79.84\pm24.12$ & $77.40\pm25.56$ & $12.76\pm25.96$ & $77.90\pm23.69$ & $80.06\pm24.39$ & $81.26\pm22.63$ & $4.55\pm5.42$ \tabularnewline
+t2f & $98.96\pm00.31$ & $90.33\pm01.58$ & $23.83\pm09.12$ & $61.40\pm04.55$ & $76.33\pm24.72$ & $79.46\pm24.52$ & $78.35\pm25.25$ & $12.10\pm21.87$ & $77.96\pm23.79$ & $79.72\pm24.65$ & $81.77\pm22.85$ & $4.78\pm5.95$\tabularnewline
+t2w & $99.44\pm00.46$ & $90.29\pm01.68$ & $24.31\pm08.09$ & $61.88\pm04.05$ & $77.28\pm23.70$ & $80.82\pm23.37$ & $78.67\pm24.37$ & $10.45\pm22.12$ & $78.79\pm22.91$ & $81.03\pm23.75$ & $81.86\pm22.35$ & $4.63\pm5.96$\tabularnewline
\end{tabular}
}
\label{tab:cav-postop-seg-results}
\end{table}

\begin{figure}[!ht]
\centering
\includegraphics[scale=0.56]{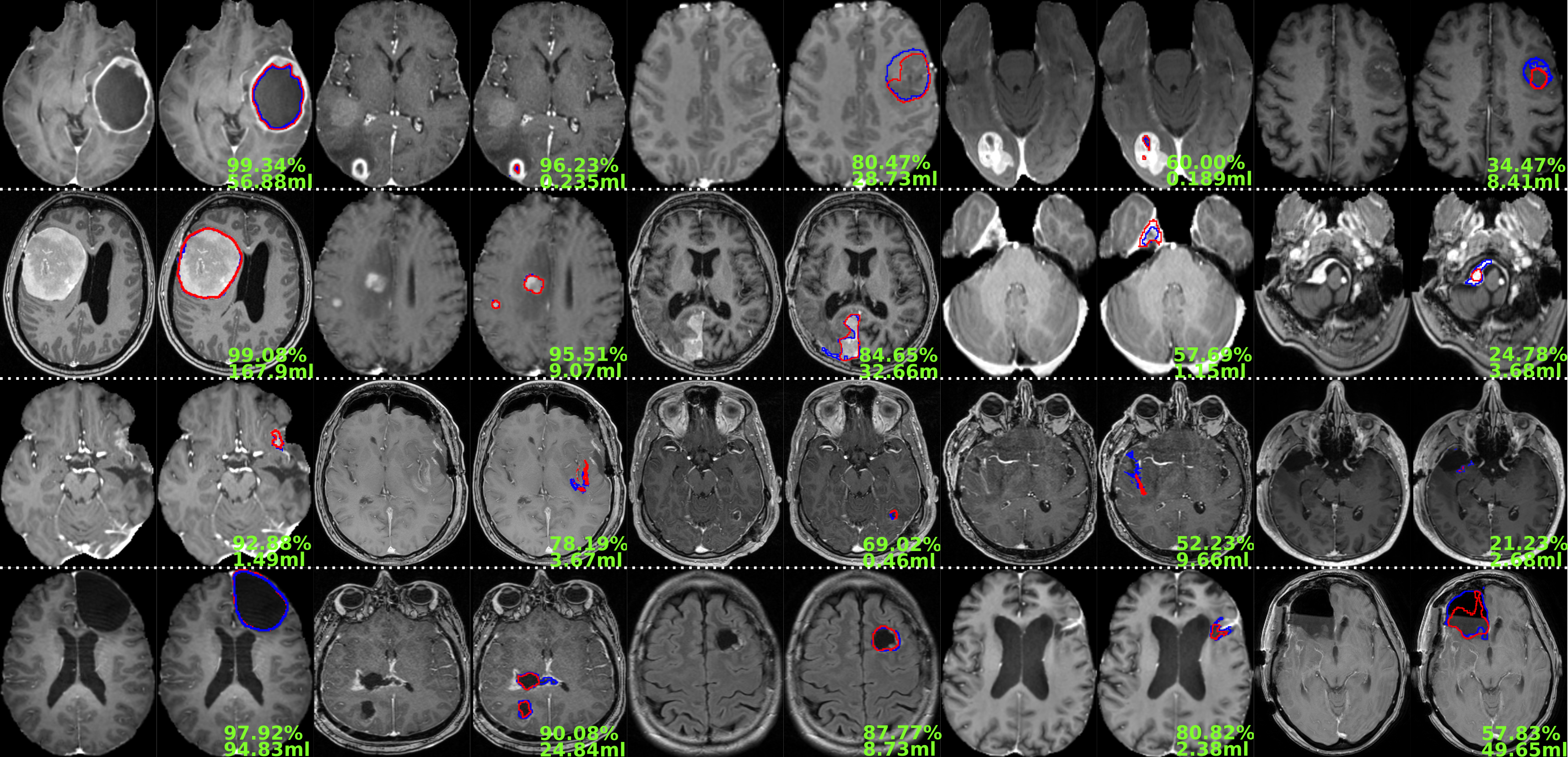}
\caption{Illustration showing the ground truth (in blue) against the produced prediction (in red) for the NETC, tumor core, residual tumor, and resection cavity from top to bottom. The resulting Dice score and total volume to segment are given in green (image best viewed digitally and in color).}
\label{fig:results-examples-illu}
\end{figure}

An overview of the models' performance for each structure of interest is provided in Fig.~\ref{fig:results-examples-illu}, showing examples of best to worse results from left to right. For large tumors with a clear contrast-enhancing rim, the NETC model segmented almost perfectly with pixel-wise Dice scores above 95\%. When the rim is not visible, potentially in presence of low-/non-contrast-enhancing tumors, the model tended to struggle. Regarding the contrast-enhancing tumor core segmentation model, clearly defined CNS tumors were almost perfectly segmented regardless of size (i.e., glioblastoma or metastasis). Identified cases of struggle often exhibit CNS tumors in unusual location (e.g., around the brain stem) not featured enough in the dataset. Next, the postoperative residual tumor model faced the challenge of segmenting small, fragmented, and not clearly confined structures. Oftentimes only part of the residual tumor was correctly segmented, omitting other smaller components around the cavity. The larger pieces of non-resected tumor, similar to the contrast-enhancing tumor core, were more easily segmented. Finally, the resection cavity segmentation model was more deficient when presented with inhomogeneous cavities displaying varying intensity levels (cf. right-most example in the last row of the figure).

\subsection{(iv) Segmentation performances analysis across cohorts and BraTS challenge benchmarking}

The comparison of the contrast-enhancing tumor core segmentation performances, against our previous baseline and the latest BraTS challenge results from 2024, is compiled in Table.~\ref{tab:seg-results-preop-benchmark}. In our previous study, a specific tumor core segmentation was trained for each tumor type, while a unified model is proposed in this article for all contrast-enhancing tumors. For both the glioma and meningioma groups, the voxel-wise performances have improved from training a unified model, with a Dice score gain of 5\% and 2\%. The results over the glioma category are quite meaningful to interpret, given the relative high magnitude of both datasets with more than $2\,000$ patients included. However, the difference in dataset size for both the meningioma and metastasis categories makes for a difficult direct comparison, as tumors' expression and variability might greatly differ.

\begin{table}[h]
\caption{Preoperative contrast-enhancing tumor core segmentation performances, compared to the latest BraTS challenge performances and our previous baseline~\cite{bouget2023raidionics}}
\adjustbox{max width=\textwidth}{
\begin{tabular}{rr||cccc||cccc||cccc}
 & & \multicolumn{4}{c||}{Patient-wise} & \multicolumn{4}{c||}{Voxel-wise} & \multicolumn{4}{c}{Object-wise} \tabularnewline
 \hline
Fold & \# Samples & Recall & Precision & Specificity & bAcc & Dice & Recall & Precision & HD95 & Dice & Recall & Precision & HD95\tabularnewline
GLI & 3700 & $99.40\pm00.34$ & $99.86\pm00.12$ & $50.15\pm44.72$ & $74.78\pm22.42$ & $89.44\pm12.40$ & $90.16\pm12.20$ & $90.77\pm12.17$ & $4.29\pm10.19$ & $88.73\pm12.41$ & $89.30\pm13.01$ & $90.40\pm12.52$ & $2.87\pm3.75$ \tabularnewline
BraTS leaderboard & 219 & - & - & - & - & - & - & - & - & $87.0$ & - & - & -\tabularnewline
Baseline & 2134 & - & - & - & - & $86.63\pm12.41$ & $98.32\pm01.10$ & $95.35\pm02.13$ & - & - & $85.06\pm07.87$ & $91.96\pm03.33$ & -\tabularnewline
\hline
MEN & 2278 & $96.92\pm00.46$ & $99.95\pm00.09$ & $94.75\pm10.00$ & $95.84\pm04.97$ & $88.83\pm13.62$ & $88.49\pm15.46$ & $91.34\pm11.46$ & $7.93\pm20.55$ & $89.26\pm12.90$ & $88.52\pm15.52$ & $92.26\pm09.88$ & $2.42\pm3.91$\tabularnewline
BraTS leaderboard & 70 & - & - & - & - & - & - & - & - & $84.9$ & - & - & -\tabularnewline
Baseline & 719 & - & - & - & - & $86.62\pm14.54$ & $94.90\pm03.29$ & $92.32\pm04.32$ & - & - & $88.36\pm03.33$ & $84.32\pm07.99$ & -\tabularnewline
\hline
MET & 1214 & $99.74\pm00.20$ & $98.99\pm01.01$ & $75.99\pm23.59$ & $87.86\pm11.73$ & $85.68\pm14.30$ & $84.28\pm16.13$ & $89.61\pm12.51$ & $10.36\pm21.72$ & $84.94\pm18.38$ & $86.37\pm14.27$ & $90.87\pm11.49$ & $1.48\pm1.30$\tabularnewline
BraTS leaderboard & 88 & - & - & - & - & - & - & - & - & $81.0$ & - & - & -\tabularnewline
Baseline & 396 & - & - & - & - & $88.58\pm14.08$ & $97.73\pm02.09$ & $95.61\pm03.46$ & - & - & $82.94\pm05.13$ & $92.67\pm05.58$ & -\tabularnewline
\end{tabular}
}
\label{tab:seg-results-preop-benchmark}
\end{table}

Segmentation performances of postoperative contrast-enhancing residual tumors across cohorts, with more than 100 samples, have been reported in Table~\ref{tab:postop-seg-across-cohorts}. For fair comparison with the BraTS challenge, performances are reported for the model trained using all four MR sequences as input. A complete report over all cohorts is available in the supplementary material (Table.S8). A direct comparison with the results from the BraTS 2024 challenge is not possible since the test set is not publicly available. As a result, the latest official leaderboard results over the validation set (available on Synapse), were used in the table for benchmarking purposes.
An average lesion-wise Dice score of $76.3\%$ was obtained from the best-performing team for the enhancing tissue category, computed over 188 samples. From our model, averaged over the 1316 available samples from the BraTS challenge, an object-wise Dice score of $79.7\%$ is obtained, hence on-par with the state-of-the-art.
However, large performance variations can be noticed between the main dataset's cohorts. The pixel-wise Dice score decreases to 61.6\% for the STO cohort and goes further down to 47\% over the SUH cohort. A similar trend is visible across the different cohorts for the patient-wise and object-wise metrics. The performance drop seems to be highly correlated with the average residual tumor volume over each cohort (cf. Table.S3) going from 15\,ml over the BraTS cohort down to 3.7\,ml for the SUH cohort.

\begin{table}[h]
\caption{Contrast-enhancing tumor segmentation performances for cohorts with at least 100 patients, compared to the latest BraTS challenge performances, using all four input sequences.}
\adjustbox{max width=\textwidth}{
\begin{tabular}{rr||cccc||cccc||cccc}
 & & \multicolumn{4}{c||}{Patient-wise} & \multicolumn{4}{c||}{Voxel-wise} & \multicolumn{4}{c}{Object-wise} \tabularnewline
Fold & \# Samples & Recall & Precision & Specificity & bAcc & Dice & Recall & Precision & HD95 & Dice & Recall & Precision & HD95\tabularnewline
STO & 407 & $94.69\pm01.36$ & $71.49\pm03.96$ & $59.75\pm09.61$ & $77.22\pm04.28$ & $61.64\pm23.10$ & $64.41\pm27.70$ & $68.10\pm24.26$ & $10.77\pm14.14$ & $61.86\pm23.16$ & $67.37\pm27.23$ & $73.52\pm20.85$ & $6.44\pm6.24$\tabularnewline
SUH & 199 & $77.18\pm06.07$ & $87.91\pm08.23$ & $84.51\pm09.20$ & $80.84\pm04.68$ & $47.16\pm30.60$ & $50.90\pm34.51$ & $51.02\pm33.20$ & $9.01\pm12.05$ & $49.67\pm30.73$ & $54.19\pm34.84$ & $74.37\pm23.22$ & $4.14\pm5.11$ \tabularnewline
BraTS & 1316 & $98.36\pm01.36$ & $89.42\pm03.67$ & $77.88\pm05.55$ & $88.12\pm02.91$ & $80.27\pm19.36$ & $81.71\pm19.13$ & $82.44\pm21.08$ & $5.93\pm10.91$ & $79.77\pm20.51$ & $82.19\pm20.48$ & $83.42\pm20.04$ & $3.25\pm4.42$\tabularnewline
\hline
BraTS leaderboard & 188 & - & - & - & - & - & - & - & - & $76.30$ & - & - & \tabularnewline
\end{tabular}
}
\label{tab:postop-seg-across-cohorts}
\end{table}

For the resection cavity, performances over cohorts with at least 100 samples are reported in Table.~\ref{tab:postop-seg-cavity-across-cohorts}, using the same method as described above to benchmark against the BraTS challenge 2024. A complete report over all cohorts is available in the supplementary material (Table.S9-S10). An average lesion-wise Dice score of $71.5\%$ was obtained from the best-performing team for the resection cavity segmentation. From our best model over the BraTS cohort, an object-wise Dice score of $76.33\%$ is obtained. The inter-cohort variability is much lower with average Dice scores around 80\%. Interestingly, as reported in Table.S4, average resection cavity volumes are relatively similar across all cohorts, at around 20\,ml.

\begin{table}[h]
\caption{Resection cavity segmentation performances for cohorts with at least 100 patients, compared to the latest BraTS challenge performances, using all four input sequences.}
\adjustbox{max width=\textwidth}{
\begin{tabular}{rr||cccc||cccc||cccc}
 & & \multicolumn{4}{c||}{Patient-wise} & \multicolumn{4}{c||}{Voxel-wise} & \multicolumn{4}{c}{Object-wise} \tabularnewline
Cohort & \# Samples & Recall & Precision & Specificity & bAcc & Dice & Recall & Precision & HD95 & Dice & Recall & Precision & HD95\tabularnewline
\hline
STO & 275 & $99.64\pm00.85$ & $100.00\pm00.00$ & $100.00\pm00.00$ & $99.82\pm00.43$ & $81.63\pm17.51$ & $84.71\pm17.94$ & $81.65\pm18.77$ & $14.55\pm33.81$ & $83.59\pm16.66$ & $84.53\pm18.16$ & $85.59\pm15.74$ & $4.39\pm5.08$ \tabularnewline
SUH & 165 & $100.00\pm00.00$ & $98.18\pm02.69$ & $58.18\pm48.99$ & $79.09\pm24.49$ & $76.02\pm22.89$ & $81.34\pm22.14$ & $76.14\pm23.77$ & $10.59\pm25.77$ & $77.97\pm21.60$ & $81.62\pm22.28$ & $79.25\pm21.56$ & $3.49\pm5.96$ \tabularnewline
BraTS & 1316 & $99.55\pm00.29$ & $86.53\pm02.06$ & $19.58\pm06.53$ & $59.57\pm03.16$ & $75.18\pm25.78$ & $79.03\pm25.22$ & $77.00\pm26.50$ & $9.09\pm14.73$ & $76.33\pm25.49$ & $79.78\pm25.40$ & $79.98\pm25.01$ & $4.92\pm6.16$\tabularnewline
\hline
BraTS leaderboard & 188 & - & - & - & - & - & - & - & $71.50$ & - & -\tabularnewline
\end{tabular}
}
\label{tab:postop-seg-cavity-across-cohorts}
\end{table}

\section{Discussion}
This study presents a comprehensive investigation into standardized postsurgical assessment reporting for central nervous system tumors. First, unified segmentation models for both preoperative contrast-enhancing and non-enhancing tumor core structures were introduced, using the Attention U-Net architecture. Second, postoperative segmentation models for contrast-enhancing residual tumor and resection cavity were developed, using the same Attention U-Net architecture and variations in MR sequence used as inputs. Finally, classification models for MR sequence identification and tumor type differentiation were explored using DenseNet. The primary contribution of this study is the achievement of state-of-the-art segmentation performance, validated against the latest BraTS challenge results. Building upon these models, an automated surgical reporting pipeline was proposed, aligned with the latest RANO 2.0 guidelines. Lastly, all models and methods have been integrated into the Raidionics software, providing an open-access and standardized reporting solution for clinical use.

The preoperative contrast-enhancing tumor core segmentation dataset includes a large and diverse patient population, covering all major CNS tumor subtypes. For contrast-enhancing tumors, the t1c MR sequence remains the most informative and critical modality. However, a key limitation in our dataset is the inconsistent availability of additional MR sequences across patients, except those from the BraTS challenge. In practice, multiple sequences are needed in order to make the best treatment and prognosis decision (i.e., t1c, t1w, and t2f MR scans). However, since all four standard MR sequences might not always be available in every treating center, developing models able to perform well using a single sequence is of high practical value. To create unified segmentation models applicable to all CNS tumor types, the tumor core and non contrast-enhancing tumor core structures were considered separately in datasets A and B. This choice reflects the significant under-representation of necrotic and cystic regions in meningiomas and metastasis, which would introduce substantial complexity and class imbalance in training a mixed unified model.
For contrast-enhancing residual tumor segmentation, incorporating multiple MR sequences becomes imperative for achieving satisfactory performances. For example, separating blood products from contrast-enhancing tumor tissue requires looking at both t1c and t1w MR scans. Furthermore, DWI and ADC images might also need to be considered as input sequences postoperatively in the future. Those two sequences would help further disambiguate misconstrued residual disease obtained from t2f sequences for non contrast-enhancing CNS tumors, or provide information for identifying infarctions. In our dataset C, as sequence availability varies, an increase in required input modalities leads to a decrease in number of usable cases. Yet, sufficient data remains to train effective models across all configurations. Finally, dataset D is an exception as it contains a good proportion of non contrast-enhancing cases, as opposed to the other datasets. The objective of creating unified models is the main rationale for mixing contrast-enhancing and non contrast-enhancing samples. A resection cavity model operating primarily on t2f MR scans is essential to enable complete surgical reporting across all major CNS tumor types, especially if t1 sequences are not available. In order to expand the training dataset for this configuration, additional cases from the BraTS challenge cohort, originally segmented in t1c MR scans then co-registered to t2f, were used.

Performances are almost perfect for the MR sequence classification task, with only 32 misclassified cases out of $8\,000$. Each erroneous classification could be linked to a problem in the acquisition of the MR scan, especially regarding motion or illumination artifacts. Even given proper classification, it remains debatable if those MR scans would be eligible for properly running a segmentation model, given such noise level. On the other hand, the tumor type classification task exhibited some limitations as indicated by the lower multiclass accuracy. The motivation to use solely the t1c MR scan as input was motivated by the content of our dataset A, where many samples do not have more than one sequence. For contrast-enhancing tumors, the size, location, and multifocality aspect are the most intuitive characteristics to consider. As such, providing the tumor core mask as input, in an anatomy-guided fashion, should help the network to predominantly focus on those regions. Given the high segmentation performance of the tumor core segmentation model, the tumor core mask should be obtained quite robustly. Providing additional MR sequences as input, such as a t2f, might prove useful in the future for training unified classification models able to operate indiscriminately over contrast-enhancing and non contrast-enhancing tumors.

For the preoperative tumor core segmentation task over contrast-enhancing CNS tumors, the unified model exhibits better performances than type-specific models, both compared to our previous results and latest BraTS challenge leaderboards. Using only the t1c MR scans as input does not appear to be a limitation. The reported performances are on-par or higher than the reported performances from the BraTS challenge using all four input MR sequences.
Postoperatively, the contrast-enhancing residual tumor segmentation models performed well, achieving an object-wise Dice score of 70\% and a HD95 of $4.32$\,mm. Among the different input configurations, the most notable finding was the 3\% increase in object-wise Dice score when the t1w MR scan was included alongside the t1c scan. Adding the remaining two sequences yielded another 3\% improvement. However, these gains must be interpreted with caution due to potential biases introduced by varying sample sizes across configurations. Specifically, a direct comparison of performance across input combinations is limited, as each was evaluated on a different subset of patients. Notably, when all four MR sequences were included, the BraTS cohort constituted 60\% of dataset C, while the next largest cohort (STO) represented only 18\%. As such, models trained with all four sequences are more likely to learn from the BraTS cohort sample distribution, hence a potential slight bias toward this cohort. Fortunately, including the t1w sequence reduced the sample size by only 30 patients, allowing for a more valid and representative comparison. Importantly, this input configuration is clinically meaningful for contrast-enhancing CNS tumors, as the t1w scan helps distinguish residual tumor from postoperative blood products.
For the resection cavity segmentation, the incremental inclusion of MR sequences as input appears to have minimal impact on performance, with stable object-wise Dice scores around 78\%. While using solely t1c MR scan as input provides good performances, the use of t2f scan only yielded much lower performance, down to 66\% Dice score. An explanation might come from the type of CNS tumor featured in the different cohorts. In dataset D, unlike in the other datasets almost exclusively featuring contrast-enhancing tumors, a higher number of non contrast-enhancing tumors are including.
Overall, the models' classification ability does not match their segmentation performance, as reflected in the patient-wise metrics. Models tend to over-segment, resulting in high recall but lower precision and specificity, which is a known trade-off in computer vision tasks. Another explanation might be the skewed positive-to-negative ratio in the training data for most cohorts, whereby nearly all patients exhibit the structure of interest to be segmented. Specific model training techniques, known to handle better large variations in sample distribution or class imbalance (e.g., bootstrapping, hard-mining), could be used. Alternatively, architectures able to perform classification and segmentation at the same time (e.g., Mask R-CNN) could help mitigate this issue. From a clinical perspective, favoring models with higher recall and lower precision may be preferable. Missing a tumor region could have more severe consequences than falsely identifying non-tumor tissue.
Analysis of inter-cohort variability reveals several factors that may influence the performance of residual tumor segmentation. Notably, as observed in the per-cohort average volumes, a substantial difference between the BraTS challenge cohort and other cohorts exists, with BraTS structures being nearly twice as large. Such volume discrepancy may partly explain the 20\% difference in Dice scores, as segmentation models tend to perform better the bigger the target. Another key distinction is the timing of postoperative imaging during patient care. While all other cohorts include only early postoperative scans, the BraTS dataset comprises multiple postoperative time points. Scans acquired several months after surgery typically exhibit reduced cavity content and resolution of surgical blood products, making residual tumor boundaries more discernible.
These differences underscore the significant variability in MR imaging across institutions and highlight the importance of diverse, multi-institutional datasets for training models with strong generalizability. The challenge of curating data balancing quality and quantity is especially pronounced for the postoperative residual tumor segmentation task. In the context of contrast-enhancing residual tumor segmentation, additional variability may stem from differences in surgical practices (e.g., intraoperative decision-making, extent of resection) and treatment strategies (e.g., surgery, chemotherapy). This interpretation is reinforced by the more consistent performance observed across cohorts for the resection cavity segmentation, suggesting that this task is less affected by clinical variability.

A direct and exact comparison with BraTS challenge performance is not feasible due to the unavailability of their test sets and potential differences in design choices. Specifically, the implementation details of metric computation can significantly impact the results, most notably in the pairing strategy to compute object-wise metrics (referred to as lesion-wise in BraTS). Nonetheless, we believe that our own implementation provides a fair and consistent assessment of model performance. A key source of variation lies in the threshold used to determine whether a model prediction is considered correct, either on a patient-wise or object-wise basis. For instance, adopting a very low Dice threshold (e.g., $0.1\%$) between ground truth and detection favors patient-wise recall but results in lower average Dice scores. Conversely, using a stricter threshold such as $50\%$ Dice aligns more closely with conventions in computer vision, yielding improved pixel-wise and object-wise performance but at the cost of reduced sensitivity. While this stricter threshold may reduce the patient-wise recall, it arguably better reflects clinical relevance through qualitative visual agreement.
For more robust statistical comparisons across models, limiting the analysis to patients with all four MR sequences would be a more rigorous choice. However, given the limited availability of all sequences in dataset A, excluding several hundred patients would significantly reduce tumor heterogeneity in the training samples and introduce a bias toward the predominantly American data from the BraTS challenge dataset.

The proposed standardized surgical report offers an almost complete quantitative depiction of the most relevant structures, for both contrast-enhancing and non contrast-enhancing tumors. The volumetric assessment of the brain, tumor core, non-enhancing tumor core, resection cavity, and FLAIR hyperintensity structures provide the needed information to perform an assessment based on the RANO 2.0 guidelines for CNS tumors. The inclusion of MR sequence and tumor type classification models in the reporting pipeline also alleviates the burden for the user. A folder containing MR acquisitions, organized preoperatively and early postoperatively, needs only to be manually provided for starting the reporting process. Preoperative and postoperative segmentation of the most relevant structures in contrast-enhancing and non contrast-enhancing CNS tumors is fully supported. A fine-grained classification into tumor sub-types or image-based biomarker prediction would be interesting to investigate. Furthermore, the inclusion of segmentation models for postoperative complications, such as hemorrhages or infarctions, also remains to be studied.

Future work could explore the development of models capable of segmenting all relevant structures simultaneously or incorporating global context refinement mechanisms. Implementing a single unified model for all structures poses challenges, likely requiring a highly curated and specific dataset. For example, structures such as NETC are not consistently present across all tumor types and are often absent in meningiomas and metastases. As a results, a unified model trained on imbalanced data underperform relative to dedicated, structure-specific models.
Additionally, a more effective strategy is also needed to resolve the ambiguity between resection cavities and NETC regions, which may exhibit similar appearances on t1c MR scans. Such issue becomes particularly problematic in reoperation scenarios, where preexisting cavities can lead to inaccurate volumetric measurements in standardized preoperative reports. Since preoperative models are not trained to distinguish such occurrences, employing a STAPLE-based fusion approach, incorporating both preoperative and postoperative segmentation masks, may improve overall accuracy.
Another promising direction involves the development of multi-class segmentation models able to accomodate variable combinations of MR sequences~\cite{eijgelaar2020robust}. Data augmentation techniques, such as randomly masking one or more sequences during training, may improve robustness. To further capitalize on the existing datasets, including patients with missing sequences would be beneficial. In this context, the use of generative diffusion models to synthetize absent MR scans also warrants exploration.
Lastly, clinical validation of the standardized report is essential. In particular, assessing the predictive value of the automatically derived measurements in relation to clinical outcomes, such as survival and quality of life, will be essential for ensuring the clinical utility and adoption of such tools.

\section*{Acknowledgements}
Data were processed in digital labs at HUNT Cloud, Norwegian University of Science and Technology, Trondheim, Norway. 

\section*{Funding}
D.B., M.G.F, I.R., and O.S. are partly funded by the Norwegian National Research Center for Minimally Invasive and Image-Guided Diagnostics and Therapy.
P.C.D.W.H and F.B were supported by an unrestricted grant of Stichting Hanarth fonds, “Machine learning for better neurosurgical decisions in patients with glioblastoma”; a grant for public-private partnerships (Amsterdam UMC PPP-grant) sponsored by the Dutch government (Ministry of Economic Affairs) through the Rijksdienst voor Ondernemend Nederland (RVO) and Topsector Life Sciences and Health (LSH), “Picturing predictions for patients with brain tumors”; a grant from the Innovative Medical Devices Initiative program, project number 10-10400-96-14003; The Netherlands Organisation for Scientific Research (NWO), 2020.027; a grant from the Dutch Cancer Society, VU2014-7113; the Anita Veldman foundation, CCA2018-2-17.
ASJ received grants from the Swedish state. Under the agreement between the Swedish government and the county councils concerning economic support of research and education of doctors (ALF-agreement ALFGBG-1006089)

\section*{Author contributions statement}
F.B, H.A, L.B, M.S.B, M.C.N, J.F, S.L.H-J, A.J.S.I, B.K, R.N.T, E.M, P.A.R, M.R, T.S, T.A, M.W, G.W, A.H.Z, A.S.J, T.R.S, P.C.D.W.H, and O.S collected and curated the data; D.B. conceived the experiment(s); D.B. conducted the experiments; D.B. analysed the results; D.B. wrote the original draft; D.B., M.G.F, O.S, and I.R reviewed and edited the manuscript; I.R and O.S acquired the funding. All authors reviewed the manuscript. 

\section*{Data statement}
Approvals were obtained from the Norwegian regional ethics committee (REK ref. 2013/1348 and 2019/510), and from the Medical Ethics Review Committee of VU University Medical Center (IRB00002991, 2014.336).

\section*{Additional information}
\textbf{Accession codes} the Raidionics environment with all related information is available at \url{https://github.com/raidionics}. More specifically, all trained models can be accessed at \url{https://github.com/raidionics/Raidionics-models/releases/tag/1.3.1}, the Raidionics software can be found at \url{https://github.com/raidionics/Raidionics}. Finally, the source code used to compute the validation metrics is available at \url{ https://github.com/dbouget/validation_metrics_computation}

\textbf{Competing interests} The authors declare no competing interests. The funders had no role in the design of the study; in the collection, analyses, or interpretation of data; in the writing of the manuscript; or in the decision to publish the results.

\bibliographystyle{unsrt} 
\bibliography{main}  

\clearpage
\appendix
\section*{Supplementary Material}

\renewcommand{\thesection}{S\arabic{section}} 
\renewcommand{\thefigure}{S\arabic{figure}}
\renewcommand{\thetable}{S\arabic{table}}
\setcounter{section}{0}
\setcounter{figure}{0}
\setcounter{table}{0}

\section*{Data}
\label{sec:Data}

\subsection*{Cohort abbreviations}
Abbreviations used for each cohort featured in one of the four datasets used in this study are listed in the following:
\begin{enumerate}
    \item BraTS - MICCAI BraTS challenges (editions 2023 and 2024)
    \item STO - St. Olavs hospital, Trondheim University Hospital, Trondheim, Norway
    \item STOP - Polyclinics affiliated to St. Olavs hospital, Trondheim University Hospital, Trondheim, Norway
    \item SUH - Sahlgrenska University Hospital, Göteborg, Sweden
    \item UCSF - University of California San Francisco Medical Center, U.S.A
    \item ETZ - St Elisabeth Hospital, Tilburg, Netherlands
    \item VUmc - Amsterdam University Medical Centers, location VU medical center, Netherlands
    \item HMC - Medical Center Haaglanden, the Hague, Netherlands
    \item HUM - Humanitas Research Hospital, Milano, Italy
    \item UMCU -  University Medical Center Utrecht, Netherlands
    \item ISALA - Isala hospital, Zwolle, Netherlands
    \item MUW - Medical University Vienna, Austria
    \item UMCG - University Medical Center Groningen, Netherlands
    \item PARIS - Hôpital Lariboisière, Paris, France
    \item SLZ - Medical Center Slotervaart, Amsterdam, Netherlands
    \item NWZ - Northwest Clinics, Alkmaar, Netherlands
    \item BOS - Brigham and Women's Hospital, Boston, USA
    \item OSL - Oslo University Hospital, Oslo, Norway
    \item BOS - Brigham and Women's Hospital, Boston, USA
\end{enumerate}

\subsection*{Dataset statistics}
As part of the following datasets, patients suffering from glioma (GLI), glioblastoma (GBM), meningioma (MEN), and metastasis (MET) have been included. For each dataset, overview tables are reporting average values per cohort and/or CNS tumor type. In addition to total number of samples and average volumes, the total number of positive samples and the positive rate are also provided. Shortened from positive-to-negative sample ratio, the measure indicates the proportion of positive samples, i.e., containing the structure of interest.

\subsubsection*{Dataset A - Contrast-enhancing tumor core}
For dataset A, a detailed cohort-wise and tumor type wise description is presented in Table.~\ref{tab:dataset-tc-detailed}. 
The BraTS challenge is by far the biggest contributor across all tumor types, with the St. Olavs hospital placing second. Average volumes are consistent across all cohorts for corresponding tumor types. The biggest CNS tumors being glioblastomas with a volume larger than $30\,ml$ on average. It can be noticed that meningioma patients referred to surgery had much larger tumors on average than patients followed at the outpatient clinic, with a $18\,ml$ difference. A few contrast-enhancing tumors were not formally identified for the STO cohort and were labelled as \textit{Others}. In dataset A, only 61 samples out of 7212 do not contain any contrast-enhancing tumor to segment.

\begin{table}[!h]
\centering
\caption{Detailed overview of the preoperative tumor core segmentation dataset (dataset A), per cohort and tumor type.}
\adjustbox{max width=\textwidth}{
\begin{tabular}{lcc|cccc|}
Cohort & Type & Samples & Positives & Positive rate (\%) & Volume (ml)\tabularnewline
\hline
BraTS & GLI & $1663$ & $1656$ & $99.57$ & $37.87\pm32.67$\tabularnewline
BraTS & MEN & $1500$ & $1494$ & $99.60$ & $21.12\pm29.98$\tabularnewline
BraTS & MET & $817$ & $777$ & $95.10$ & $09.17\pm12.73$\tabularnewline
STO & GBM & $596$ & $596$ & $100$ & $32.93\pm32.02$\tabularnewline
STOP & MEN & $442$ & $440$ & $99.54$ & $12.34\pm21.72$\tabularnewline
STO & MEN & $336$ & $336$ & $100$ & $30.81\pm34.70$\tabularnewline
STO & MET & $332$ & $332$ & $100$ & $19.79\pm18.57$\tabularnewline
SUH & GBM & $251$ & $251$ & $100$ & $35.46\pm28.19$\tabularnewline
UMCU & GBM & $171$ & $171$ & $100$ & $36.82\pm27.96$\tabularnewline
ETZ & GBM & $153$ & $153$ & $100$ & $36.83\pm28.86$\tabularnewline
UCSF & GBM & $134$ & $133$ & $99.25$ & $28.69\pm26.83$\tabularnewline
HMC & GBM & $103$ & $103$ & $100$ & $42.17\pm29.45$\tabularnewline
VUmc & GBM & $97$ & $97$ & $100$ & $31.52\pm24.01$\tabularnewline
UMCG & GBM & $86$ & $86$ & $100$ & $32.26\pm27.86$\tabularnewline
MUW & GBM & $83$ & $83$ & $100$ & $34.35\pm29.33$\tabularnewline
HUM & GBM & $75$ & $75$ & $100$ & $26.71\pm22.26$\tabularnewline
PARIS & GBM & $74$ & $72$ & $97.29$ & $32.44\pm23.67$\tabularnewline
ISALA & GBM & $72$ & $72$ & $100$ & $37.62\pm31.18$\tabularnewline
OUS & MET & $67$ & $64$ & $95.52$ & $06.27\pm07.18$\tabularnewline
SLZ & GBM & $49$ & $49$ & $100$ & $36.87\pm28.38$\tabularnewline
NWZ & GBM & $38$ & $38$ & $100$ & $26.47\pm23.76$\tabularnewline
STO & Others & $28$ & $28$ & $100$ & $21.20\pm20.04$\tabularnewline
\end{tabular}
}
\label{tab:dataset-tc-detailed}
\end{table}

\newpage
\subsubsection*{Dataset B - Non-enhancing tumor core}
For dataset B, a detailed cohort-wise and tumor type wise description is presented in Table.~\ref{tab:dataset-necro-detailed}. For the glioma category, a distinction is made between preoperative and postoperative acquisitions, while only preoperative data are available for the other two tumor types.
The BraTS challenge is the only contributor, and the NETC structure is predominantly featured in preoperative gliomas, with a 95\% occurrence rate and 15\,ml average volume. For the other groups, NETC is only featured with a 40\% positive rate and a much lower average volume around 3.5\,ml.

\begin{table}[!ht]
\centering
\caption{Detailed overview of the NETC segmentation dataset (dataset B). Preoperative data are indicated with preop. and postoperative data with postop.}
\adjustbox{max width=\textwidth}{
\begin{tabular}{lc|cccc|}
Cohort & Type & Samples & Positives & Positive rate (\%) & Volume (ml)\tabularnewline
\hline
BraTS & GLI preop. & $1294$ & $1224$ & $94.59$ & $15.61\pm21.94$\tabularnewline
BraTS & GLI postop. & $1316$ & $492$ & $37.38$ & $04.56\pm06.92$\tabularnewline
BraTS & MEN & $1000$ & $343$ & $34.30$ & $03.18\pm08.87$ \tabularnewline
BraTS & MET & $817$ & $448$ & $54.83$ & $03.64\pm06.77$\tabularnewline
\end{tabular}
}
\label{tab:dataset-necro-detailed}
\end{table}

\subsubsection*{Dataset C - Residual tumor (enhancing tissue)}
For dataset C, a detailed cohort-wise description is presented in Table.~\ref{tab:dataset-et-detailed}. The rate of positive to negative samples is highly varying from cohort to cohort, from 34\% at the lowest up to 74\%. On average, residual tumor volumes are similar across all but one cohort, ranging from 3\,ml to 6\,ml. The BraTS challenge represents an exception where the samples exhibit residual tumor with an average volume of 15\,ml.

\begin{table}[!ht]
\centering
\caption{Detailed overview of the contrast-enhancing residual tumor dataset per cohort (dataset C).}
\adjustbox{max width=\textwidth}{
\begin{tabular}{c|cccc|}
Cohort & Samples & Positives & Positive rate (\%) & Volume (ml)\tabularnewline
\hline
BraTS & $1316$ & $875$ & $66.48$ & $15.25\pm21.41$\tabularnewline
STO & $421$ & $220$ & $52.25$ & $6.35\pm8.18$\tabularnewline
SUH & $200$ & $119$ & $59.50$ & $3.67\pm5.72$\tabularnewline
UCSF & $109$ & $81$ & $74.31$ & $4.92\pm9.11$\tabularnewline
ETZ & $101$ & $73$ & $72.27$ & $4.94\pm5.19$\tabularnewline
VUmc & $74$ & $52$ & $70.27$ & $2.74\pm3.86$\tabularnewline
HMC & $67$ & $43$ & $64.18$ & $5.06\pm6.38$\tabularnewline
HUM & $56$ & $35$ & $62.50$ & $3.70\pm3.90$\tabularnewline
UMCU & $54$ & $21$ & $38.88$ & $4.27\pm4.91$\tabularnewline 
ISALA & $52$ & $18$ & $34.61$ & $3.86\pm6.75$\tabularnewline
MUW & $51$ & $35$ & $68.62$ & $3.91\pm5.94$\tabularnewline
UMCG & $48$ & $29$ & $60.41$ & $4.52\pm5.30$\tabularnewline
PARIS & $43$ & $31$ & $72.09$ & $4.37\pm4.82$\tabularnewline
SLZ & $28$ & $21$ & $75.00$ & $3.84\pm5.41$\tabularnewline
NWZ & $27$ & $21$ & $77.77$ & $3.76\pm4.69$\tabularnewline
\end{tabular}
}
\label{tab:dataset-et-detailed}
\end{table}

\newpage
\subsubsection*{Dataset D - Resection cavity}
For dataset D, a detailed cohort-wise description is presented in Table.~\ref{tab:dataset-rc-overview}. For all cohorts but BraTS, a resection cavity is present for each sample with an almost 100\% positive rate. For the BraTS cohort, around 18\% of all samples do not feature a resection cavity. For the three most-populated cohorts, an average resection cavity volume of 16\,ml is measured. Some of the less populated cohorts do exhibit average volumes up to 30\,ml. The BOS cohort is comprised solely of non-contrast-enhancing tumors, a mixture of contrast-enhancing (322) and non-contrast-enhancing (74) tumors is present in the STO cohort, and the BraTS cohort also includes a few non-contrast-enhancing tumors.

\begin{table}[!ht]
\centering
\caption{Detailed overview of the resection cavity segmentation dataset per cohort (dataset D).}
\adjustbox{max width=\textwidth}{
\begin{tabular}{c|cccc|}
Cohort & Samples & Positives & Positive rate (\%) & Volume (ml)\tabularnewline
\hline
BraTS & $1316$ & $1092$ & $82.97$ & $16.71\pm22.09$\tabularnewline
STO & $396$ & $396$ & $100$ & $17.38\pm13.59$\tabularnewline
BOS & $236$ & $235$ & $99.57$ & $15.07\pm16.29$\tabularnewline
SUH & $167$ & $164$ & $98.20$ & $22.09\pm18.61$\tabularnewline
ETZ & $57$ & $57$ & $100$ & $21.72\pm15.62$\tabularnewline
ISALA & $16$ & $16$ & $100$ & $31.21\pm20.75$\tabularnewline
VUMC & $16$ & $16$ & $100$ & $20.97\pm12.05$\tabularnewline
MUW & $15$ & $15$ & $100$ & $20.97\pm14.68$\tabularnewline
UMCG & $12$ & $12$ & $100$ & $17.97\pm11.92$\tabularnewline
UCSF & $10$ & $10$ & $100$ & $29.70\pm16.96$\tabularnewline
HMC & $9$ & $9$ & $100$ & $20.97\pm18.89$\tabularnewline
PARIS & $9$ & $9$ & $100$ & $30.20\pm11.49$\tabularnewline
UMCU & $9$ & $9$ & $100$ & $26.04\pm13.28$\tabularnewline
HUM & $4$ & $4$ & $100$ & $30.87\pm24.63$\tabularnewline
SLZ & $4$ & $4$ & $100$ & $28.39\pm25.12$\tabularnewline
NWZ & $3$ & $3$ & $100$ & $14.49\pm04.18$\tabularnewline
\end{tabular}
}
\label{tab:dataset-rc-overview}
\end{table}

\clearpage
\section*{Results}

\subsection*{Preoperative tumor core and NETC segmentation performances}

A detailed cohort-wise performance summary is presented in Table.~\ref{tab:tc-cohortwise-results} for the preoperative contrast-enhancing tumor core segmentation, over t1c MR scans from dataset A. The classification ability of the model is perfect for half the cohorts and not dropping under 97\% for the other half, in terms of recall and precision. Since the positive rate in dataset A lies around 99\%, the specificity and balanced accuracy values are drastically lowered when a few negative patients are misclassified.
Regarding the segmentation performance, the object-wise Dice scores span from 86\% up to 91\% for all samples where the patient was given surgical treatment, indicating a strong ability to generalize. For the STOP cohort, the average object-wise Dice score was slightly lower at 84\%, for patients followed at the outpatient clinic, with one of the lowest average tumor volume.

\begin{table}[!h]
\caption{Preoperative contrast-enhancing tumor core segmentation performances, over t1c MR scans from dataset A, per cohort.}
\adjustbox{max width=\textwidth}{
\begin{tabular}{rr||cccc||ccc||ccc}
& & \multicolumn{4}{c||}{Patient-wise} & \multicolumn{3}{c||}{Voxel-wise} & \multicolumn{3}{c}{Object-wise} \tabularnewline
Cohort & \# Samples & Recall & Precision & Specificity & bAcc & Dice & Recall & Precision & Dice & Recall & Precision\tabularnewline
ETZ & 153 & $100.00\pm00.00$ & $100.00\pm00.00$ & $100.00\pm00.00$ & $100.00\pm00.00$ & $89.98\pm09.76$ & $92.20\pm10.52$ & $88.63\pm10.62$ & $88.18\pm10.90$ & $91.17\pm11.51$ & $86.96\pm11.78$\tabularnewline
HMC & 103 & $100.00\pm00.00$ & $100.00\pm00.00$ & $100.00\pm00.00$ & $100.00\pm00.00$ & $90.25\pm08.58$ & $92.43\pm09.20$ & $88.56\pm09.61$ & $89.62\pm08.78$ & $91.92\pm10.04$ & $88.53\pm07.24$\tabularnewline
HUM & 75 & $100.00\pm00.00$ & $100.00\pm00.00$ & $100.00\pm00.00$ & $100.00\pm00.00$ & $90.33\pm06.41$ & $93.93\pm04.89$ & $87.59\pm09.25$ & $89.35\pm06.68$ & $93.44\pm05.67$ & $86.61\pm09.79$\tabularnewline
ISALA & 72 & $100.00\pm00.00$ & $100.00\pm00.00$ & $100.00\pm00.00$ & $100.00\pm00.00$ & $90.60\pm08.25$ & $88.39\pm10.59$ & $93.60\pm06.57$ & $90.77\pm07.22$ & $88.65\pm08.65$ & $93.65\pm07.39$\tabularnewline
BraTS & 3980 & $98.94\pm00.13$ & $99.59\pm00.09$ & $40.82\pm16.85$ & $69.88\pm08.41$ & $86.93\pm17.91$ & $85.99\pm18.79$ & $90.26\pm16.63$ & $86.43\pm19.46$ & $86.57\pm18.46$ & $91.59\pm14.21$\tabularnewline
MUW & 83 & $100.00\pm00.00$ & $100.00\pm00.00$ & $100.00\pm00.00$ & $100.00\pm00.00$ & $89.13\pm10.52$ & $94.04\pm06.74$ & $86.25\pm11.97$ & $88.62\pm06.99$ & $92.82\pm08.66$ & $86.22\pm08.92$\tabularnewline
NWZ & 38 & $100.00\pm00.00$ & $100.00\pm00.00$ & $100.00\pm00.00$ & $100.00\pm00.00$ & $87.40\pm11.87$ & $91.75\pm11.47$ & $84.71\pm13.24$ & $88.26\pm08.14$ & $93.50\pm06.62$ & $84.99\pm11.59$\tabularnewline
OUS & 67 & $98.51\pm05.00$ & $98.51\pm04.00$ & $85.07\pm40.00$ & $91.79\pm19.53$ & $90.18\pm05.68$ & $88.41\pm06.59$ & $92.56\pm07.09$ & $87.31\pm08.08$ & $84.83\pm09.72$ & $92.01\pm06.50$\tabularnewline
PARIS & 74 & $100.00\pm00.00$ & $98.65\pm02.35$ & $77.03\pm40.00$ & $88.51\pm20.00$ & $89.12\pm07.18$ & $91.94\pm08.33$ & $87.26\pm08.29$ & $87.93\pm12.35$ & $92.05\pm09.16$ & $86.36\pm12.41$\tabularnewline
SLZ & 49 & $100.00\pm00.00$ & $100.00\pm00.00$ & $100.00\pm00.00$ & $100.00\pm00.00$ & $89.28\pm07.68$ & $90.66\pm07.50$ & $88.75\pm10.48$ & $86.28\pm10.65$ & $87.45\pm10.54$ & $88.01\pm10.66$\tabularnewline
STO & 1283 & $99.45\pm00.40$ & $100.00\pm00.00$ & $100.00\pm00.00$ & $99.73\pm00.20$ & $88.06\pm16.72$ & $89.79\pm15.47$ & $88.99\pm16.98$ & $88.09\pm16.12$ & $89.22\pm15.51$ & $90.30\pm14.47$\tabularnewline
STOP & 442 & $97.29\pm01.21$ & $100.00\pm00.00$ & $100.00\pm00.00$ & $98.64\pm00.60$ & $84.03\pm20.87$ & $84.54\pm22.02$ & $85.30\pm20.55$ & $84.21\pm20.63$ & $84.49\pm21.76$ & $88.97\pm13.93$\tabularnewline
SUH & 251 & $99.60\pm00.74$ & $100.00\pm00.00$ & $100.00\pm00.00$ & $99.80\pm00.37$ & $88.43\pm12.60$ & $87.33\pm14.45$ & $91.11\pm11.66$ & $87.65\pm13.23$ & $86.37\pm15.19$ & $91.48\pm10.17$\tabularnewline
UCSF & 134 & $100.00\pm00.00$ & $99.25\pm02.00$ & $85.07\pm40.00$ & $92.54\pm20.00$ & $89.59\pm07.91$ & $89.50\pm10.06$ & $90.75\pm08.01$ & $88.72\pm07.86$ & $88.80\pm09.89$ & $90.18\pm08.45$\tabularnewline
UMCG & 86 & $100.00\pm00.00$ & $100.00\pm00.00$ & $100.00\pm00.00$ & $100.00\pm00.00$ & $88.61\pm09.90$ & $92.65\pm10.94$ & $86.53\pm11.04$ & $87.48\pm10.49$ & $91.18\pm12.59$ & $86.27\pm10.15$\tabularnewline
UMCU & 171 & $99.42\pm01.03$ & $100.00\pm00.00$ & $100.00\pm00.00$ & $99.71\pm00.51$ & $87.17\pm09.55$ & $90.60\pm09.59$ & $85.07\pm12.07$ & $86.11\pm09.81$ & $89.64\pm10.57$ & $84.85\pm10.51$\tabularnewline
VUmc & 97 & $100.00\pm00.00$ & $100.00\pm00.00$ & $100.00\pm00.00$ & $100.00\pm00.00$ & $89.81\pm09.32$ & $91.40\pm08.86$ & $89.58\pm12.29$ & $88.32\pm10.22$ & $89.76\pm10.16$ & $89.08\pm12.50$\tabularnewline
\end{tabular}
}
\label{tab:tc-cohortwise-results}
\end{table}

\begin{figure}[!h]
\centering
\includegraphics[scale=0.5]{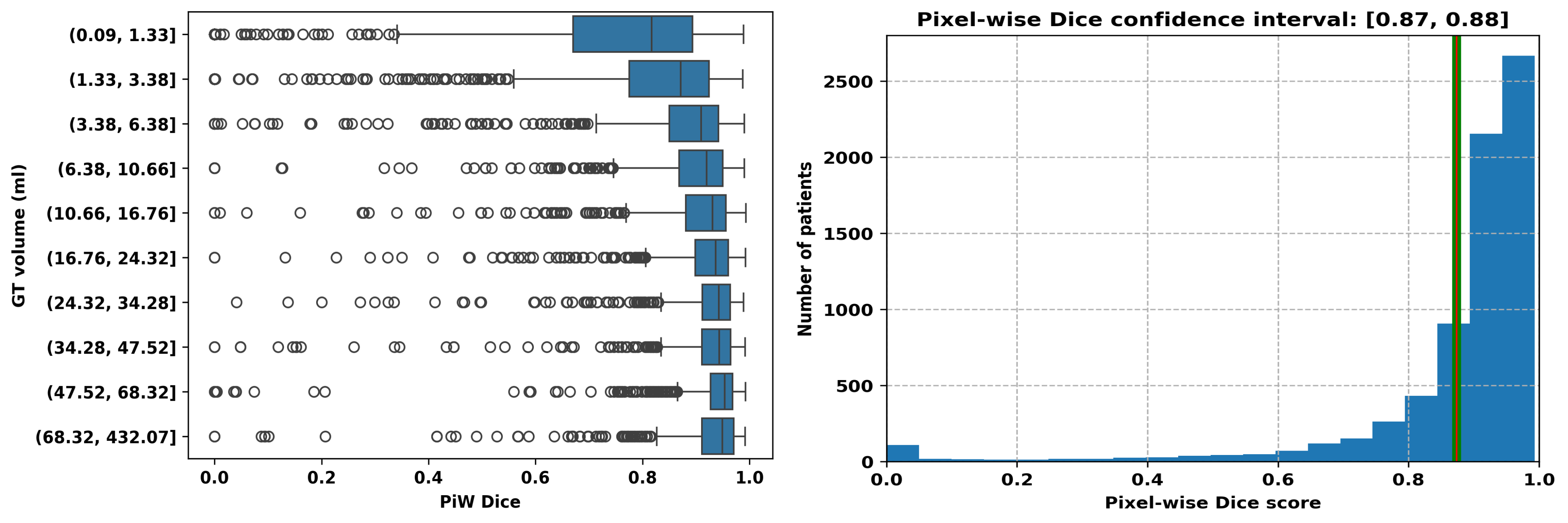}
\caption{Boxplot showing the voxel-wise Dice against tumor core volume for ten equally populated bins (to the left) and voxel-wise confidence intervals (to the right) for all preoperative contrast-enhancing tumor core positive samples.}
\label{fig:tumorcore-seg-boxplot-ci}
\end{figure}

There is a relationship between model performance and tumor size, the smaller the structure to detect the harder the task. The relationship is further outlined by the equally-populated boxplots showing the relation between tumor volume and voxel-wise Dice score (cf. left-hand side illustration in Fig.~\ref{fig:tumorcore-seg-boxplot-ci}). The average voxel-wise Dice score is above 90\% for tumors bigger than 6\,ml, above 80\% for tumors larger than 1.5\,ml, and finally at 72\% for tumor smaller than 1.5\,ml.
A total of 94 cases were completely missed, meaning a voxel-wise Dice score of 0\%, of which 49 exhibit a tumor core smaller than 1\,ml. The confidence interval is closely around the reported average voxel-wise Dice score of 87.3\%, indicating a low variability and high confidence in the estimated average (cf. right-hand side illustration in Fig.~\ref{fig:tumorcore-seg-boxplot-ci}).

\begin{table}[!b]
\caption{NETC segmentation performance, per tumor type featured in the BraTS cohort, for the model using t1c, t1w, and t2f as inputs.}
\adjustbox{max width=\textwidth}{
\begin{tabular}{rr||cccc||ccc||ccc}
& & \multicolumn{4}{c||}{Patient-wise} & \multicolumn{3}{c||}{Voxel-wise} & \multicolumn{3}{c}{Object-wise} \tabularnewline
Type & \# Samples & Recall & Precision & Specificity & bAcc & Dice & Recall & Precision & Dice & Recall & Precision\tabularnewline
GBM & 2610 & $97.56\pm01.59$ & $85.85\pm03.51$ & $68.83\pm08.48$ & $83.19\pm03.62$ & $73.06\pm26.18$ & $79.18\pm27.21$ & $75.77\pm24.08$ & $73.64\pm26.14$ & $81.45\pm26.95$ & $76.17\pm24.11$\tabularnewline
Meningioma & 1000 & $76.27\pm02.21$ & $62.92\pm02.55$ & $76.50\pm02.53$ & $76.38\pm00.68$ & $56.14\pm27.58$ & $60.84\pm31.91$ & $63.12\pm26.71$ & $56.18\pm32.90$ & $71.46\pm35.38$ & $69.81\pm29.15$\tabularnewline
Metastasis & 817 & $96.62\pm01.32$ & $78.09\pm06.63$ & $67.29\pm08.33$ & $81.95\pm04.21$ & $69.79\pm21.32$ & $74.34\pm25.35$ & $72.56\pm20.25$ & $68.09\pm26.11$ & $77.47\pm25.59$ & $75.15\pm21.52$\tabularnewline
\end{tabular}
}
\label{tab:necro-typewise-results}
\end{table}

Regarding the NETC structure, a detailed performance summary is presented in Table.~\ref{tab:necro-typewise-results}, for the segmentation model using t1c, t1w, and t2f MR sequences as inputs. Unsurprisingly, the model performs best on the glioma tumor type, where the non-enhancing tumor core structure is typically present with a relatively high volume. On the contrary, the model struggles in meningiomas as the NETC structure is rare and small (cf. Table~\ref{tab:dataset-necro-detailed}), reaching only 56\% Dice score. For equally small NETC structures, the Dice score is 12\% higher over the metastasis group compared to the meningioma one. The positive rate in the former is 20\% higher than in the latter, hence providing more positive sample during training.

In order to further investigate the model capabilities, a detailed summary is presented in Table.~\ref{tab:necro-timepointwise-results} illustrating the relationship between segmentation performances and stage of care. The highest Dice score, above 70\%, was obtained over patient data acquired before surgery, where the tumor is potentially at its largest hence containing the most NETC. On the other hand, Dice scores over patient data acquired after surgery (i.e., early post-op and consecutive post-operative checks) are barely nearing 52\%. Since, hopefully, the largest part of the tumor core has been removed during surgery, very little NETC is left to segment, making the task more difficult. In addition, potential confusion may arise between the post-operative NETC structure and resection cavity, both being visually very similar.

\begin{table}[!t]
\caption{NETC segmentation performances per stage of care, for the model using t1c, t1w, and t2f as inputs.}
\adjustbox{max width=\textwidth}{
\begin{tabular}{rr||cccc||ccc||ccc}
 & & \multicolumn{4}{c||}{Patient-wise} & \multicolumn{3}{c||}{Voxel-wise} & \multicolumn{3}{c}{Object-wise} \tabularnewline
Fold & \# Samples & Recall & Precision & Specificity & bAcc & Dice & Recall & Precision & Dice & Recall & Precision\tabularnewline
Pre-op & 2459 & $94.61\pm01.36$ & $87.51\pm00.78$ & $71.24\pm02.78$ & $82.93\pm01.02$ & $70.50\pm29.75$ & $74.92\pm31.85$ & $72.69\pm28.20$ & $70.02\pm30.84$ & $79.71\pm30.36$ & $77.49\pm24.57$\tabularnewline
Follow-up & 652 & $96.54\pm01.71$ & $77.74\pm08.49$ & $69.62\pm09.41$ & $83.08\pm04.82$ & $67.03\pm24.86$ & $71.19\pm28.96$ & $70.49\pm24.15$ & $65.17\pm29.10$ & $74.86\pm28.94$ & $74.94\pm23.48$\tabularnewline
Early post-op & 613 & $93.46\pm05.93$ & $64.09\pm06.79$ & $68.97\pm09.25$ & $81.21\pm02.00$ & $52.91\pm31.23$ & $62.94\pm34.41$ & $59.65\pm31.43$ & $55.86\pm33.18$ & $65.50\pm36.14$ & $67.77\pm31.02$\tabularnewline
Post-op 1 & 484 & $97.40\pm02.81$ & $71.86\pm05.68$ & $76.26\pm06.58$ & $86.83\pm02.84$ & $54.11\pm29.53$ & $63.24\pm33.35$ & $60.32\pm29.67$ & $57.94\pm31.33$ & $68.75\pm34.52$ & $65.67\pm30.21$\tabularnewline
Post-op 2 & 109 & $98.17\pm04.00$ & $69.58\pm14.47$ & $66.93\pm12.36$ & $82.55\pm06.26$ & $51.72\pm30.17$ & $65.85\pm35.05$ & $58.12\pm30.20$ & $59.60\pm29.01$ & $72.44\pm33.65$ & $66.58\pm27.21$\tabularnewline
Post-op 3 & 70 & $100.00\pm00.00$ & $61.87\pm22.68$ & $72.84\pm19.07$ & $86.42\pm09.54$ & $52.90\pm31.66$ & $58.89\pm35.98$ & $66.92\pm26.98$ & $59.92\pm25.07$ & $66.08\pm27.69$ & $68.80\pm29.10$\tabularnewline
Post-op 4 & 40 & $88.75\pm20.00$ & $80.33\pm27.45$ & $78.00\pm32.00$ & $83.37\pm15.36$ & $51.07\pm27.71$ & $59.41\pm32.17$ & $56.43\pm30.41$ & $51.89\pm27.09$ & $63.52\pm32.06$ & $61.98\pm25.58$\tabularnewline
\end{tabular}
}
\label{tab:necro-timepointwise-results}
\end{table}

\begin{figure}[!h]
\centering
\includegraphics[scale=0.5]{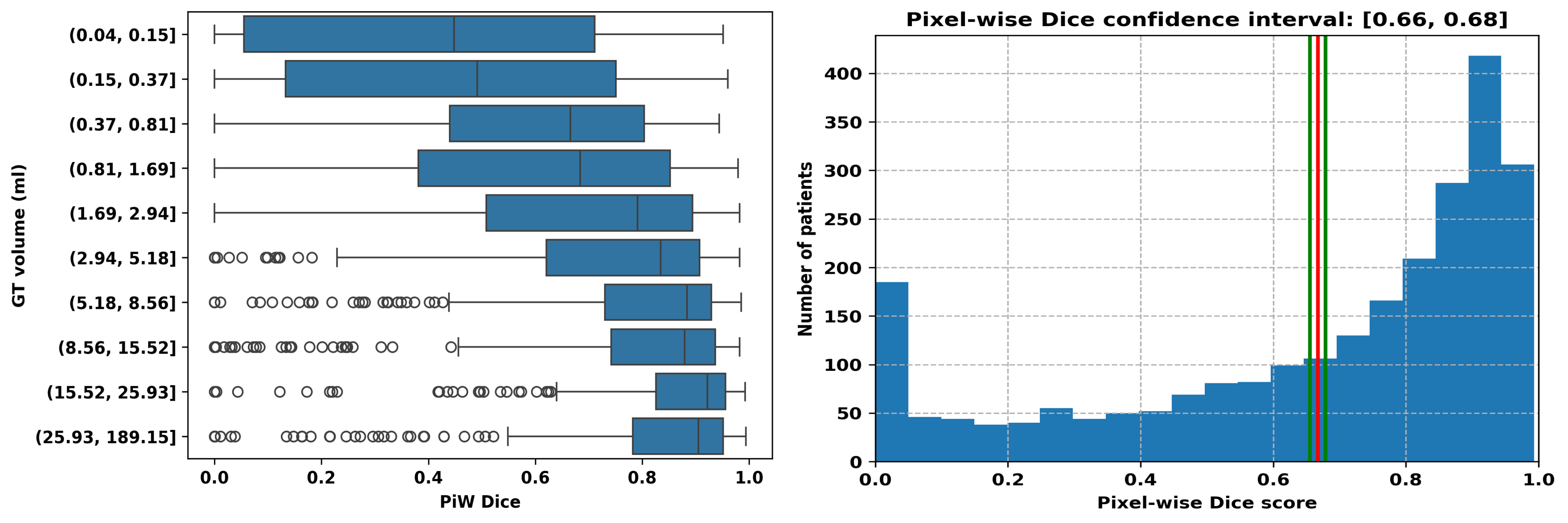}
\caption{Boxplot showing the voxel-wise Dice against NETC volume for ten equally populated bins (to the left) and voxel-wise confidence intervals (to the right) for all positive NETC samples.}
\label{fig:necrosis-seg-boxplot-ci}
\end{figure}

The average voxel-wise Dice score is above 80\% for NETC bigger than 10\,ml, around 75\% for NETC larger than 3\,ml, and finally at 50\% for NETC smaller than 1\,ml (cf. left-hand side illustration in Fig.~\ref{fig:necrosis-seg-boxplot-ci}).
A total of 137 cases were completely missed, of which 108 exhibit a NETC structure smaller than 1\,ml. The confidence interval is closely around the reported average voxel-wise Dice score of 66.7\% (cf. right-hand side illustration in Fig.~\ref{fig:necrosis-seg-boxplot-ci}).

\subsection*{Postoperative segmentation performances}

A detailed cohort-wise performances summary is presented in Table.~\ref{tab:et-postop-seg-results} for the postoperative contrast-enhancing residual tumor segmentation, using all four MR sequences as input. The cohorts not featuring any patient with all four sequences available were dropped from the table. The best voxel-wise Dice score was obtained for the BraTS cohort with 76\%, followed by the STO cohort at 60\%, and then most of the other cohorts at 45\%. From the very limited number of patients with all MR sequences in many cohorts (i.e., less than 15 patients), it is difficult to judge whether the model struggled to generalize over out-of-distribution cases or not. On average, the residual tumor volumes are much larger over the BraTS cohort (15\,ml), larger over the STO cohort (6\,ml), and then around 4\,ml for the other cohorts. The trend is clearly visible here also whereby a model struggles to segment smaller structures.

\begin{table}[!t]
\caption{Overall segmentation performance summary for postoperative contrast-enhancing residual tumor, using all four input sequences from dataset C.}
\adjustbox{max width=\textwidth}{
\begin{tabular}{rr||cccc||ccc||ccc}
 & & \multicolumn{4}{c||}{Patient-wise} & \multicolumn{3}{c||}{Voxel-wise} & \multicolumn{3}{c}{Object-wise} \tabularnewline
Fold & \# Samples & Recall & Precision & Specificity & bAcc & Dice & Recall & Precision & Dice & Recall & Precision\tabularnewline
HMC & 40 & $93.00\pm16.00$ & $75.15\pm05.05$ & $37.08\pm29.33$ & $65.04\pm14.85$ & $29.24\pm20.74$ & $49.14\pm32.22$ & $33.53\pm25.70$ & $35.68\pm22.84$ & $56.21\pm33.51$ & $44.74\pm30.93$\tabularnewline
HUM & 5 & $60.00\pm47.14$ & $60.00\pm47.14$ & $80.00\pm23.57$ & $70.00\pm35.36$ & $19.07\pm21.35$ & $39.26\pm41.66$ & $18.46\pm19.65$ & $19.08\pm21.35$ & $39.26\pm41.66$ & $18.46\pm19.65$\tabularnewline
BraTS & 1316 & $97.48\pm00.88$ & $84.71\pm02.89$ & $59.63\pm05.73$ & $78.56\pm02.80$ & $77.45\pm23.30$ & $79.42\pm22.86$ & $79.42\pm25.18$ & $76.44\pm25.27$ & $82.41\pm21.03$ & $82.38\pm22.18$\tabularnewline
MUW & 44 & $97.73\pm06.67$ & $77.23\pm20.84$ & $56.82\pm29.77$ & $77.27\pm13.15$ & $51.81\pm22.55$ & $58.58\pm26.03$ & $53.92\pm23.97$ & $49.58\pm17.99$ & $59.60\pm23.56$ & $55.30\pm20.77$\tabularnewline
PARIS & 38 & $100.00\pm00.00$ & $75.66\pm15.33$ & $27.63\pm40.00$ & $63.82\pm20.00$ & $49.15\pm17.48$ & $64.75\pm23.30$ & $46.53\pm19.05$ & $49.00\pm20.09$ & $70.70\pm25.01$ & $46.02\pm20.54$\tabularnewline
STO & 407 & $95.89\pm00.85$ & $66.88\pm03.84$ & $45.12\pm10.30$ & $70.51\pm05.51$ & $59.83\pm24.52$ & $63.76\pm28.19$ & $66.00\pm26.35$ & $60.20\pm24.58$ & $67.48\pm27.83$ & $72.11\pm22.85$\tabularnewline
SUH & 199 & $75.13\pm05.81$ & $89.69\pm08.84$ & $83.83\pm13.46$ & $79.48\pm05.97$ & $44.01\pm31.74$ & $47.85\pm35.45$ & $47.55\pm34.46$ & $46.23\pm32.13$ & $51.22\pm36.24$ & $72.80\pm26.50$\tabularnewline
UCSF & 6 & $58.33\pm41.46$ & $58.33\pm41.46$ & $50.00\pm43.30$ & $54.17\pm32.48$ & $23.74\pm36.93$ & $27.10\pm42.04$ & $21.15\pm33.01$ & $24.61\pm38.16$ & $27.10\pm42.04$ & $47.54\pm40.23$\tabularnewline
UMCG & 43 & $97.67\pm08.00$ & $74.78\pm13.71$ & $32.44\pm34.41$ & $65.06\pm13.78$ & $43.86\pm27.11$ & $58.92\pm33.02$ & $43.36\pm25.75$ & $45.50\pm26.87$ & $65.05\pm31.98$ & $50.08\pm27.82$\tabularnewline
UMCU & 46 & $95.96\pm05.71$ & $62.73\pm08.33$ & $52.03\pm12.00$ & $74.00\pm04.92$ & $41.34\pm21.97$ & $66.30\pm28.15$ & $37.47\pm24.29$ & $42.13\pm24.06$ & $69.69\pm29.47$ & $47.45\pm28.59$\tabularnewline
VUmc & 72 & $96.30\pm08.89$ & $83.28\pm05.58$ & $29.68\pm23.13$ & $62.99\pm09.00$ & $49.07\pm27.38$ & $59.66\pm31.65$ & $50.15\pm29.39$ & $51.72\pm28.61$ & $64.08\pm31.86$ & $58.06\pm30.59$\tabularnewline
\end{tabular}
}
\label{tab:et-postop-seg-results}
\end{table}

\begin{figure}[!h]
\centering
\includegraphics[scale=0.5]{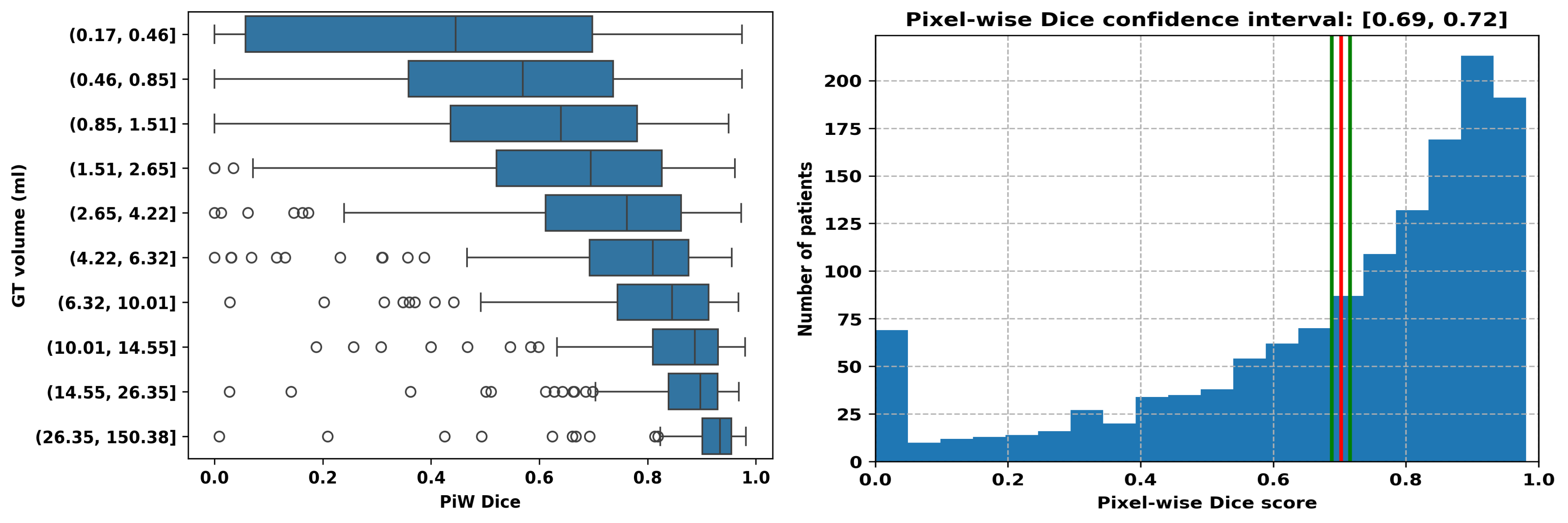}
\caption{Boxplot showing the voxel-wise Dice against residual tumor volume for ten equally populated bins (to the left) and voxel-wise confidence intervals (to the right) for all positive residual tumor samples.}
\label{fig:residualtumor-seg-boxplot-ci}
\end{figure}

The trend is outlined by the equally-populated boxplots showing the relation between structure volume and voxel-wise Dice score (cf. left-hand side illustration in Fig.~\ref{fig:residualtumor-seg-boxplot-ci}). An average voxel-wise Dice score of 80\% was obtained for residual tumor bigger than 2\,ml, 61\% for residual tumor larger than 1.0\,ml, and finally at 48\% for residual tumor smaller than 1\,ml. Out of the $1\,375$ positive cases with all four inputs available, 61 were completely missed by the model. The confidence interval is closely around the reported average voxel-wise Dice score of 70.2\% (cf. right-hand side illustration in Fig.~\ref{fig:residualtumor-seg-boxplot-ci})

Finally, regarding the resection cavity structure, a detailed cohort-wise performances summary is presented in Table.~\ref{tab:cav-seg-cohortwise-results}, for the model using all four MR sequences as input. Across the board, the object-wise Dice scores are relatively stable, ranging from 77\% (SUH cohort) up to 94\% (PARIS cohort). For the two cohorts featuring some cases without any visible resection cavity, the patient-wise specificity and balanced accuracy are negatively impacted as the model has been trained with a bias towards expecting to segment a cavity.

\begin{table}[!h]
\caption{Resection cavity segmentation performances cohort-wise, using all four MR sequences as input.}
\adjustbox{max width=\textwidth}{
\begin{tabular}{rr||cccc||ccc||ccc}
& & \multicolumn{4}{c||}{Patient-wise} & \multicolumn{3}{c||}{Voxel-wise} & \multicolumn{3}{c}{Object-wise} \tabularnewline
Type & \# Samples & Recall & Precision & Specificity & bAcc & Dice & Recall & Precision & Dice & Recall & Precision\tabularnewline
HMC & 8 & $100.00\pm00.00$ & $100.00\pm00.00$ & $100.00\pm00.00$ & $100.00\pm00.00$ & $77.12\pm26.19$ & $75.76\pm25.99$ & $80.38\pm29.82$ & $77.12\pm26.19$ & $75.76\pm25.99$ & $80.38\pm29.82$\tabularnewline
ISALA & 5 & $100.00\pm00.00$ & $100.00\pm00.00$ & $100.00\pm00.00$ & $100.00\pm00.00$ & $88.16\pm07.11$ & $93.17\pm04.86$ & $84.82\pm12.62$ & $89.89\pm04.15$ & $93.17\pm04.86$ & $87.42\pm07.66$\tabularnewline
BraTS & 1316 & $99.55\pm00.29$ & $86.53\pm02.06$ & $19.58\pm06.53$ & $59.57\pm03.16$ & $75.18\pm25.78$ & $79.03\pm25.22$ & $77.00\pm26.50$ & $76.33\pm25.49$ & $79.78\pm25.40$ & $79.98\pm25.01$\tabularnewline
MUW & 14 & $100.00\pm00.00$ & $100.00\pm00.00$ & $100.00\pm00.00$ & $100.00\pm00.00$ & $87.87\pm10.18$ & $87.07\pm12.98$ & $91.48\pm10.16$ & $89.83\pm08.98$ & $87.07\pm12.98$ & $94.95\pm04.81$\tabularnewline
PARIS & 9 & $100.00\pm00.00$ & $100.00\pm00.00$ & $100.00\pm00.00$ & $100.00\pm00.00$ & $93.99\pm00.94$ & $93.15\pm02.23$ & $94.95\pm02.38$ & $94.01\pm00.96$ & $93.15\pm02.23$ & $94.98\pm02.37$\tabularnewline
STO & 275 & $99.64\pm00.85$ & $100.00\pm00.00$ & $100.00\pm00.00$ & $99.82\pm00.43$ & $81.63\pm17.51$ & $84.71\pm17.94$ & $81.65\pm18.77$ & $83.59\pm16.66$ & $84.53\pm18.16$ & $85.59\pm15.74$\tabularnewline
SUH & 165 & $100.00\pm00.00$ & $98.18\pm02.69$ & $58.18\pm48.99$ & $79.09\pm24.49$ & $76.02\pm22.89$ & $81.34\pm22.14$ & $76.14\pm23.77$ & $77.97\pm21.60$ & $81.62\pm22.28$ & $79.25\pm21.56$\tabularnewline
UCSF & 2 & $100.00\pm00.00$ & $100.00\pm00.00$ & $100.00\pm00.00$ & $100.00\pm00.00$ & $91.73\pm07.04$ & $90.83\pm06.03$ & $92.66\pm08.10$ & $91.76\pm07.01$ & $90.83\pm06.03$ & $92.71\pm08.02$\tabularnewline
UMCG & 11 & $100.00\pm00.00$ & $100.00\pm00.00$ & $100.00\pm00.00$ & $100.00\pm00.00$ & $88.42\pm09.18$ & $87.38\pm12.72$ & $91.46\pm07.01$ & $88.73\pm12.04$ & $86.30\pm15.86$ & $93.80\pm03.47$\tabularnewline
UMCU & 6 & $100.00\pm00.00$ & $100.00\pm00.00$ & $100.00\pm00.00$ & $100.00\pm00.00$ & $82.61\pm11.08$ & $86.41\pm18.21$ & $83.59\pm13.25$ & $83.12\pm10.93$ & $86.55\pm18.18$ & $84.21\pm12.69$\tabularnewline
VUmc & 15 & $100.00\pm00.00$ & $100.00\pm00.00$ & $100.00\pm00.00$ & $100.00\pm00.00$ & $87.86\pm12.07$ & $88.10\pm10.64$ & $88.92\pm14.47$ & $90.62\pm05.80$ & $88.10\pm10.64$ & $94.59\pm02.83$\tabularnewline
\end{tabular}
}
\label{tab:cav-seg-cohortwise-results}
\end{table}

Specifically for the resection cavity segmentation model using the t2f as single input, a detailed cohort-wise performances summary is presented in Table.~\ref{tab:cav-seg-flair-cohortwise-results}. Surprisingly, the object-wise Dice score obtained over the BraTS cohort is only of 58\%, under the average score of 66.5\%, uncharacteristic from the previous results obtained this cohort both for the resection cavity and other structures. Conversely, the object-wise Dice score reaches at least 70\% for the other two main cohorts (i.e., BOS and STO). An explanation might come from the type of CNS tumor featured in the different cohorts. The BOS cohort features only non contrast-enhancing tumors, while the BraTS cohort features predominantly contrast-enhancing tumors.

\begin{table}[!t]
\caption{Resection cavity segmentation performances cohort-wise, using t2f only as input.}
\adjustbox{max width=\textwidth}{
\begin{tabular}{rr||cccc||ccc||ccc}
 & & \multicolumn{4}{c||}{Patient-wise} & \multicolumn{3}{c||}{Voxel-wise} & \multicolumn{3}{c}{Object-wise} \tabularnewline
Fold & \# Samples & Recall & Precision & Specificity & bAcc & Dice & Recall & Precision & Dice & Recall & Precision\tabularnewline
BOS & 236 & $99.15\pm01.05$ & $99.58\pm00.91$ & $81.36\pm40.00$ & $90.25\pm19.79$ & $69.97\pm22.29$ & $76.13\pm25.15$ & $68.45\pm21.90$ & $70.05\pm22.95$ & $75.77\pm26.06$ & $71.07\pm21.28$\tabularnewline
BraTS & 1316 & $99.18\pm01.05$ & $85.18\pm02.79$ & $07.93\pm03.05$ & $53.55\pm01.41$ & $56.76\pm31.90$ & $62.01\pm33.27$ & $59.04\pm32.21$ & $58.16\pm32.02$ & $61.99\pm33.90$ & $61.88\pm33.02$\tabularnewline
STO & 70 & $100.00\pm00.00$ & $100.00\pm00.00$ & $100.00\pm00.00$ & $100.00\pm00.00$ & $76.82\pm16.34$ & $83.95\pm19.37$ & $74.88\pm16.55$ & $79.43\pm15.03$ & $85.37\pm17.88$ & $77.52\pm16.43$\tabularnewline
UCSF & 4 & $100.00\pm00.00$ & $100.00\pm00.00$ & $100.00\pm00.00$ & $100.00\pm00.00$ & $62.27\pm11.54$ & $96.47\pm00.58$ & $47.02\pm12.53$ & $62.34\pm11.47$ & $96.47\pm00.58$ & $47.09\pm12.46$\tabularnewline
\end{tabular}
}
\label{tab:cav-seg-flair-cohortwise-results}
\end{table}

\begin{figure}[!h]
\centering
\includegraphics[scale=0.5]{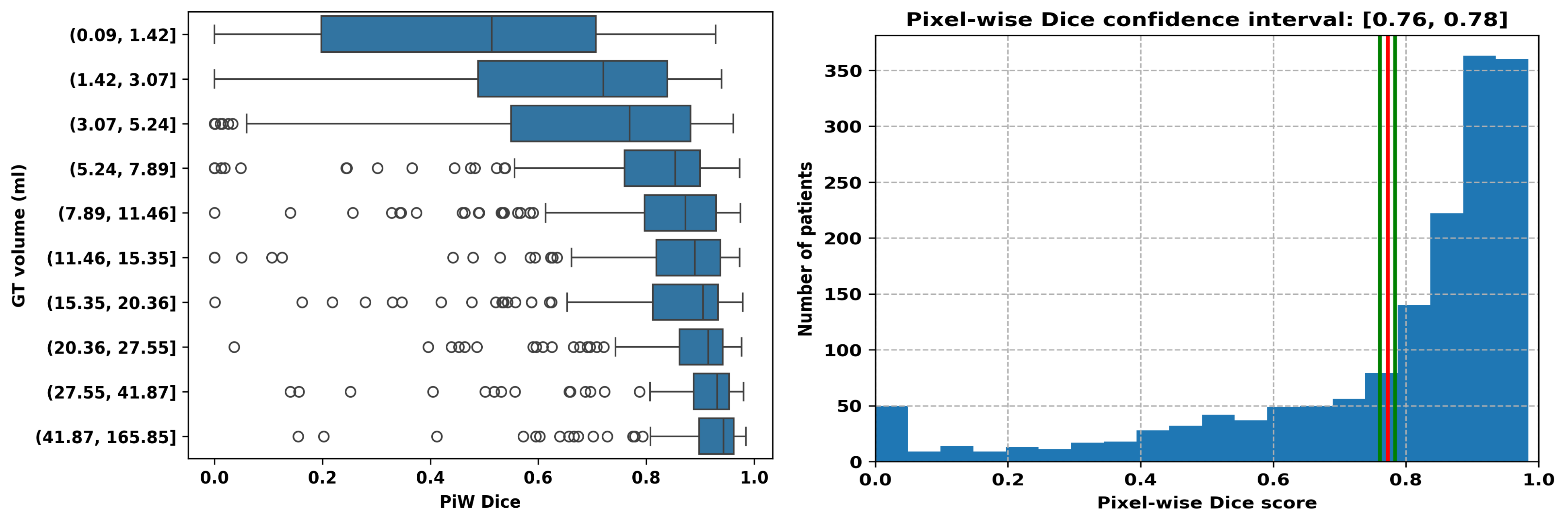}
\caption{Boxplot showing the voxel-wise Dice against resection cavity volume for ten equally populated bins (to the left) and voxel-wise confidence intervals (to the right) for all positive resection cavity samples (with the model using all four MR scans as input).}
\label{fig:cavity-seg-boxplot-ci}
\end{figure}

A similar trend as outlined before can be witnessed in the boxplots showing the relation between structure volume and voxel-wise Dice score (cf. left-hand side illustration in Fig.~\ref{fig:cavity-seg-boxplot-ci}). An average voxel-wise Dice score of 85\% was obtained for resection cavities bigger than 5\,ml, and 59\% for resection cavities smaller than 5\,ml. Out of all the positive cases having all four inputs available, 34 were completely missed by the model.
The confidence interval is closely around the reported average voxel-wise Dice score of 77.2\% (cf. right-hand side illustration in Fig.~\ref{fig:cavity-seg-boxplot-ci})

\end{document}